%% file: main.tex
\documentclass[10pt,twocolumn,letterpaper]{article}
\usepackage[pagenumbers]{cvpr}
\input{preamble}
\definecolor{cvprblue}{rgb}{0.21,0.49,0.74}
\usepackage[pagebackref,breaklinks,colorlinks,allcolors=cvprblue]{hyperref}
\title{EVA-RSI: Recursive Self-Improvement of Video Evaluators\\for Judging Temporal Quality}
\author{Yu Yuan$^{1,2}$\thanks{Work done during an internship at Roblox.}\quad
Ruizhou Ding$^{1}$\quad Shaoheng Fang$^{1,3}$\quad Stanley H. Chan$^{2}$\quad Zhuo Chen$^{1}$\\[6pt]
$^{1}$Roblox\qquad $^{2}$Purdue University\qquad $^{3}$University of Texas at Austin}
\date{}
\hypersetup{
 pdftitle={EVA-RSI: Recursive Self-Improvement of Video Evaluators for Judging Temporal Quality},
 pdfauthor={Yu Yuan, Ruizhou Ding, Shaoheng Fang, Stanley H. Chan, Zhuo Chen}
}

\begin{document}
\maketitle
\input{fig/teaser}
\input{sec/0_abstract}
\input{sec/1_intro}
\input{sec/2_related}
\input{sec/3_method}
\input{sec/4_experiments}

\input{sec/5_downstream}
\input{sec/5_conclusion}

\section*{Acknowledgments}
We thank Tinghui Zhou, Jiapeng Tang, Haiyue Yu, Charles Shang, and Kris Kitani for helpful discussions.

\section*{LLM Usage Statement}
Generative AI tools (Codex, Claude Code) assisted with language editing and figure preparation. The authors designed these protocols, audited their outputs, verified the reported results, and take full responsibility for the paper.

{
    \small
    \bibliographystyle{ieeenat_fullname}
    \bibliography{main}
}

\clearpage
\input{appendix}

\end{document}

%% file: preamble.tex
\usepackage[T1]{fontenc}

\usepackage{cuted} 

\usepackage{currfile} 

\usepackage{caption} 
\usepackage{pgfplots}
\pgfplotsset{compat=1.17}
\usetikzlibrary{arrows.meta}
\definecolor{cOurs}{HTML}{2F6DB5}
\definecolor{cOursMid}{HTML}{6F9FD8}
\definecolor{cOursLight}{HTML}{A7C4E8}
\definecolor{cVR}{HTML}{E08A3C}
\definecolor{cBase}{HTML}{9AA3AE}
\definecolor{cGood}{HTML}{3C8D6E}
\definecolor{cBad}{HTML}{C8494B}
\definecolor{cInk}{HTML}{3A4250}
\usepackage{colortbl}
\definecolor{cOursRow}{HTML}{EAF1FA}
\newcommand{\oursrow}{\rowcolor{cOursRow}}
\newcommand{\gain}[2]{\cellcolor{cOurs!#1}$+#2$}
\pgfplotsset{
  paper/.style={
    axis lines=left, axis line style={-, draw=cInk!60, line width=0.4pt},
    tick style={draw=cInk!60, line width=0.4pt}, tick align=outside, major tick length=2pt,
    ymajorgrids, grid style={draw=cInk!8, line width=0.3pt},
    tick label style={font=\scriptsize, color=cInk}, label style={font=\scriptsize, color=cInk},
    title style={font=\scriptsize\bfseries, color=cInk},
    legend style={font=\tiny, draw=none, fill=white, fill opacity=0.85, text opacity=1, row sep=-1pt},
    legend cell align=left,
    every axis plot/.append style={line width=0.8pt, line join=round},
  },
}

\newcommand{\todo}[1]{{\color{red}#1}}

\usepackage{makecell}
\usepackage{pifont}
\AtBeginDocument{\captionsetup{singlelinecheck=false}}
\newcommand{\pending}[1]{\todo{[pending]}}
\newcommand{\ours}{EVA-RSI}

%% file: fig/teaser.tex
\begin{strip}
\centering
\vspace{-6mm}
\includegraphics[width=\textwidth]{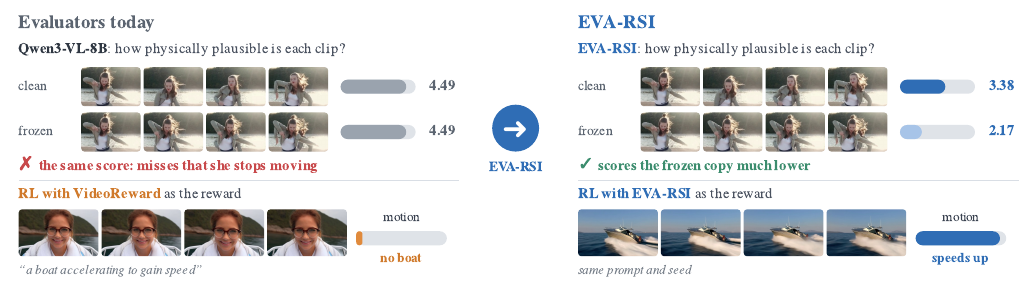}
\vspace{-5mm}
\captionsetup{hypcap=false}
\captionof{figure}{\textbf{A reward that misjudges motion trains generators that do not move.}
Top: physics ratings for a clip and for a copy in which the person freezes.
Bottom: Wan2.1 after 50 RL updates with VideoReward (visual and motion quality, no penalties) and with \ours{} as the reward.}
\captionsetup{hypcap=true}
\label{fig:overview}
\vspace{-2mm}
\end{strip}

%% file: sec/0_abstract.tex
\begin{abstract}
Video evaluators play an important role in video generation, yet they miss motion defects, limiting the quality of generated videos.
Recursive self-improvement (RSI) has made strong progress in mathematics and coding, where explicit verifiers provide reliable feedback on a single goal.
Video evaluation poses two difficulties for RSI: it has no verifier, and it must judge many dimensions at once, such as physics, smoothness and dynamics, where a gain on one can cost another.
We introduce \textbf{EVA-RSI}, a multi-agent framework for exploring RSI under these two constraints.
We build a fixed proxy exam with construction-defined targets by applying controlled changes to videos, such as temporal reversal and frame dropping.
These edits supply supervision without human annotation. A small set of human-labeled real videos anchors the exam to human judgment.
Each dimension has an LLM agent that proposes improvements. A leader agent coordinates the proposals, and a Nash bargaining rule selects updates that balance progress across dimensions.
Humans design the exam, review candidate changes and revise the curriculum between stages.
Starting from Qwen3-VL-8B, EVA-RSI improves motion evaluation, with the largest gain on smoothness, and the gains carry over to defect types never seen during the run.
On the dynamics subsets of public benchmarks, our evaluator achieves the highest observed scores among 21 tested models.
As an RL reward, it increases generated motion and earns human preference. A data-selection pipeline incorporating it also increases motion.
\end{abstract}

%% file: sec/1_intro.tex
\section{Introduction}
\label{sec:intro}

Video evaluation shapes what video generators learn.
Generators learn from evaluators, used as RL rewards~\cite{black2024ddpo,liu2025flowgrpo} and as filters for training data~\cite{blattmann2023svd,nan2025openvid}, making motion judgment central to generator training.
For example, VideoReward~\cite{liu2025videoreward}, when used as the sole RL reward for Wan2.1~\cite{wan2025}, reduces video motion after 50 updates (\cref{fig:overview}, bottom).
General vision-language models also miss motion errors: Qwen3-VL-8B~\cite{qwen3vl} assigns the same score to the original clip and an edited copy in which the person suddenly freezes (\cref{fig:overview}, top).
An evaluator that misjudges motion passes its blind spot on to the generator.

Human labels for motion are scarce, so this skill is hard to teach directly.
Recursive self-improvement (RSI) offers another route: an agent system improves a model round after round, and each round starts from the best model so far~\cite{zhang2026flexrsi,du2026rsimaster}.
In mathematics and code, models improve themselves over many rounds~\cite{zelikman2022star,guo2025deepseekr1,novikov2025alphaevolve,zhang2026dgm,zhang2026flexrsi}, because a unit test or a proof checker verifies each answer.
Video evaluation poses two difficulties for RSI.
First, it has no such verifier.
Second, it has no single goal: an evaluator must judge physics, smoothness, dynamics and more at once, and a change that helps one dimension can hurt another.
We ask: \emph{can RSI improve a video evaluator without a native verifier and across many dimensions?}

\textbf{EVA-RSI} addresses both (\cref{fig:arch}).
We build a fixed proxy exam with construction-defined targets.
We reverse a clean clip, drop frames, or make an object disappear, and assign the original clip as the pairwise target. Human-labeled real videos connect this synthetic supervision to perceptual judgments.
Each dimension has a proposer agent that studies the evaluator's errors and proposes fixes.
A leader agent coordinates the proposals, and engineer agents train new versions of the evaluator.
An acceptance rule based on Nash bargaining~\cite{nash1950bargaining} unifies the dimensions: it accepts a version only if the main dimensions improve, or hold once they saturate, and no dimension falls below its floor.
Each accepted version is a \emph{champion}.
Each round builds on the current champion, by training from it or merging it with other candidates, and its errors guide the agents' next experiments.
Between stages, humans design a better exam where the current one is too coarse, so the system combines agent search with human curriculum design.
The evaluator increases motion as an RL reward and within a data-selection pipeline; human raters prefer the RL results at update 50 (\cref{sec:downstream}).

We study EVA-RSI as an exploratory system: can a constructed exam guide useful search, can the loop improve motion judgments while retaining other skills, and do these gains benefit generation? Our contributions are:
\begin{itemize}
\item \textbf{Multi-agent RSI for video evaluation.} Two designs address the missing verifier and competing objectives: a constructed-target exam with human curriculum design, and dimension-specific agents coordinated by a leader and Nash-based acceptance (\cref{sec:method}).
\item \textbf{A stronger video evaluator.} The loop improves every main dimension over Qwen3-VL-8B, preserves the guarded skills, and transfers to unseen defects and public benchmarks (\cref{sec:exp}). Training controls and champion histories reveal how improvements accumulate and where trade-offs arise. The resulting evaluator increases generated motion as an RL reward and within a data-selection pipeline, with human preference gains at update 50 (\cref{sec:downstream}).
\end{itemize}

%% file: sec/2_related.tex
\section{Related Work}
\label{sec:related}

\noindent\textbf{Video evaluators.}
Benchmarks such as VBench~\cite{huang2024vbench}, EvalCrafter~\cite{liu2024evalcrafter} and T2V-CompBench~\cite{sun2025t2vcompbench} use a separate tool for each quality dimension.
For example, they use optical flow to measure dynamic degree.
Evaluators trained on human scores or preferences include VideoScore~\cite{he2024videoscore}, VideoPhy~\cite{bansal2025videophy,bansal2025videophy2}, VisionReward~\cite{xu2024visionreward}, LiFT~\cite{wang2024lift}, T2VQA~\cite{kou2024t2vqa}, UnifiedReward~\cite{wang2025unifiedreward}, VideoScore2~\cite{he2025videoscore2} and VideoReward~\cite{liu2025videoreward}.
VideoReward gives scores for visual quality (VQ), motion quality (MQ) and text alignment (TA).
Vision-language models are also zero-shot or trained judges~\cite{zheng2023judging,qwen3vl,xiong2024llavacritic}.
Self-Taught Evaluators~\cite{wang2024selftaught} train a text judge over iterations on synthetic pairs with known answers; we construct proxy targets for video motion and let agents search the whole training recipe.
Video understanding benchmarks~\cite{li2024mvbench,liu2024tempcompass,fu2025videomme,cai2024temporalbench,zhang2024vinoground,cores2024tvbench} show that these models miss small temporal details, and motion and physics benchmarks~\cite{wang2025motioncritic,ling2025vmbench,meng2024phygenbench,motamed2025physicsiq} find the same for generated videos.
Quality assessment models learn from human opinion scores~\cite{wu2022fastvqa,wu2023dover,wu2024qalign} and from synthetic distortions~\cite{liu2017rankiqa,lin2019kadid,wu2024qinstruct,you2024depictqa}.
Video-language models also learn temporal order from reversed or shuffled negative examples~\cite{bagad2023testoftime}.
In all of these evaluators, people choose the training recipe.
In our work, agents propose the recipes, and edits and human labels decide which recipe survives.

\noindent\textbf{Recursive self-improvement.}
Language models improve from their own outputs with self-training~\cite{zelikman2022star,huang2023selfimprove,gulcehre2023rest}, self-feedback~\cite{madaan2023selfrefine,yuan2024selfrewarding} and RL with verifiable rewards~\cite{lambert2024tulu3,guo2025deepseekr1}.
LLM agents also improve programs and agents.
They optimize prompts~\cite{yang2024opro}, evolve code~\cite{romera2024funsearch,novikov2025alphaevolve,lange2026shinka}, design or rewrite agents~\cite{zelikman2024stop,hu2025adas,zhang2026dgm,zheng2026dreamrsi} and conduct machine learning research~\cite{lu2024aiscientist,chan2025mlebench}, including autonomous post-training~\cite{du2026rsimaster} and data engineering~\cite{du2026datamaster}.
Flex-RSI~\cite{zhang2026flexrsi} tests such loops on vision and robotics tasks, and it keeps a change only if the score on held-out data increases.
These loops need a task with a reliable score, such as a unit test, a proof checker or a held-out metric.
Self-rewarding models~\cite{yuan2024selfrewarding} judge their own outputs.
The judge and model then evolve together without a fixed reference.
We apply RSI to a video evaluator, where the score itself is not reliable, and keep the verifier outside the loop.

%% file: sec/3_method.tex
\section{Method}
\label{sec:method}

\input{fig/arch}

EVA-RSI improves a video evaluator in rounds (\cref{fig:arch}).
In each round, agents propose and train new candidate evaluators (\cref{sec:loop}).
A frozen checker scores each candidate on questions with constructed answers (\cref{sec:verifier}).
Then an acceptance rule, frozen within each stage, decides whether the candidate replaces the current \emph{champion} (\cref{sec:nash}).
Each new champion anchors training in the next round.

Three choices shape the design.
First, the verifier comes before the agents: targets and scoring are fixed within each stage, making candidate scores comparable.
Second, agents control the evaluator's data, recipe, architecture and inference; the verifier remains frozen.
Third, a candidate wins only through balanced progress, so no step can buy a gain on one dimension with a large loss on another.
Humans design the curriculum and review candidate changes (\cref{sec:human}).

\subsection{Constructed answers}
\label{sec:verifier}

The checker uses fixed construction-defined targets and human labels.
The questions cover nine dimensions in four groups.

\noindent\textbf{Injected defects} (physics, consistency, smoothness, quality).
We pair each clean clip with a copy containing one defect.
For physics, the clip plays backward, or an object freezes, disappears or is duplicated.
For consistency, we swap the background, insert a scene cut or change the colors suddenly.
For smoothness, we freeze, drop, shake or flicker frames.
For quality, we blur, add noise or compress the clip heavily.
Each defect also has subtle versions, for example a freeze of 2--3 frames.
The clean clip is the assigned pairwise target.
We ask the question as a pair (``which video is better?'') and as two separate 1--5 ratings with construction-assigned targets. These targets encode the edit intervention; human labels assess perceived quality on real videos.

\noindent\textbf{Motion from optical flow} (dynamics).
We pair clips that have clearly different optical flow, and assign the higher-flow clip as the dynamics target.

\noindent\textbf{Real generated videos} (real; stage 2).
We pair generated videos that have clearly different human scores in VideoPhy-2~\cite{bansal2025videophy2} and VideoFeedback~\cite{he2024videoscore}.
These questions check that gains on injected defects transfer to real generated videos.

\noindent\textbf{Guards against forgetting} (semantics, human preference, general; general in stage 2).
These questions use caption matching, expert preference pairs from VideoGen-RewardBench~\cite{liu2025videoreward}, and video question answering from MVBench~\cite{li2024mvbench} and TempCompass~\cite{liu2024tempcompass}.

Scores combine pairwise correctness (ties: $0.5$), normalized rating agreement $1-|\mathbb{E}[r]-t|/4$ with target $t$, and multiple-choice accuracy. We average within families, then dimensions; the main mean averages five dimensions.

Two development banks, each with about 2.9k questions, drive the loop.
A hidden test bank uses other clips and prompts and adds defect types that are not in the development banks.
It shares no questions, source clips or prompts with the development banks, and we use it to score champions, not to accept candidates.
Training uses a separate split from the exam and public evaluation data.

\subsection{A team of agents}
\label{sec:loop}

Coding agents based on language models run the loop (\cref{fig:arch}) in four roles: proposers, a leader, engineers and a reviewer.
Humans design the fifth part, the verifier (\cref{sec:human}).

\noindent\textbf{Proposers.}
Each dimension has a proposer, which speaks for the player of that dimension in the bargaining game (\cref{sec:nash}): five for the main dimensions and four guardians for semantics, quality, human preference and general understanding (four and three in stage 1).
Each proposer reviews the champion's scores, its weakest question families and past candidates. It proposes up to two experiments and flags risks in other proposals.

\noindent\textbf{Leader.}
The leader picks, merges or changes proposals, one for each free GPU slot, to maximize the expected Nash product.
Each round covers multiple dimensions, including a bold idea and one protecting the guardian closest to its floor. The leader specifies each experiment.

\noindent\textbf{Engineers.}
An engineer turns one specification into code and a training recipe from the current champion, with at most 1.5 hours of training on four GPUs.
If the job fails, it reads the log and repairs the candidate once.

\noindent\textbf{Shared memory.}
A ledger records every candidate: its idea, the proposing role, its development score on each dimension, the decision and the reason for a rejection.
Proposers and the leader read it every round, so they combine ideas that worked and drop ideas that failed.
This memory, together with training from the champion, makes the search recursive: each round starts from what earlier rounds found.

A fixed program trains and scores each candidate and applies the acceptance rule.

\noindent\textbf{Reviewer.}
The reviewer reads the code changes of each candidate that the rule accepts and can veto it for rule violations, such as training on bank questions or benchmark data.
A candidate becomes champion only if it passes both the rule and the review.

\subsection{Nash bargaining acceptance}
\label{sec:nash}

We treat each dimension $i$ as a player with utility $u_i = s_i - d_i$.
Here, $s_i$ is the score of the candidate.
The disagreement point $d_i$ is the score of the start model minus a small tolerance.
A candidate replaces the champion when it meets three conditions.
First, each dimension stays above its floor ($u_i > 0$).
Second, the main dimensions improve on average, with no large drop on any of them; once they saturate, they stay within $0.005$ (\cref{sec:app_rule}).
Third, the weighted log Nash product increases:
\begin{equation}
\textstyle\sum_i w_i \log(s_i - d_i) > \sum_i w_i \log(s_i^{\text{champ}} - d_i).
\label{eq:nash}
\end{equation}
The main dimensions are physics, consistency, smoothness, dynamics and real, and each has weight 1.
The guard dimensions have smaller weights.
The rule suits competing dimensions for two reasons.
First, it penalizes proximity to a dimension's floor, because $\log u_i$ falls quickly as $u_i \to 0$.
Second, the Nash comparison is invariant to positive rescaling of each utility. Floors and drop limits separately specify score tolerances.
Family guards additionally bound score drops within each defect family (\cref{sec:app_rule}).

\subsection{Human supervision and search controls}
\label{sec:human}

\begin{table}[t]
\centering
\footnotesize
\setlength{\tabcolsep}{3pt}
\resizebox{\linewidth}{!}{%
\begin{tabular}{ll}
\toprule
Safeguard & Targeted failure \\
\midrule
Two banks, 64 question types, both orders, ratings & a shortcut on one bank or format \\
Floors and family guards & trading one skill for the mean \\
Hidden test bank & overfitting the development banks \\
12 public benchmarks & overfitting the exam as a whole \\
Frozen scorer and rule, diff filter & editing the judge, training on banks \\
Reviewer agent, human inspection & violations the program misses \\
\bottomrule
\end{tabular}}
\caption{\textbf{Controls on evaluator search.}}
\label{tab:safeguards}
\end{table}

Humans construct the exam, score the start model on each family, and replace saturated or noisy families.
During a run, they review decisions and reasons for rejection. Between stages, they revise the exam.
When a stage-1 candidate raised smoothness by giving up the time-reversal family, they added guards for single families; when the stage-1 gains did not reach real generated videos, they added real-video questions (\cref{sec:exp_insight}).
At the end, they judge the generated videos in a blind study (\cref{sec:downstream}).

The search uses four layers of controls (\cref{tab:safeguards}).
\textbf{A large and diverse exam}: 64 question types over nine dimensions, in two development banks from disjoint sources.
Each pair is asked in both orders and as separate ratings to measure order sensitivity and single-video judgments. Floors and family guards constrain score drops.
\textbf{A hidden test bank} with other clips, prompts and defect types scores only champions and never decides acceptance. Agents have no access to its questions. One stage-2 prompt names a weakness observed on the stage-1 test bank. Excluding the 36 questions on that weakness raises the score from $0.906$ to $0.916$.
\textbf{Public benchmarks}: 12 benchmarks assess transfer beyond the exam. MVBench supplies separate videos for the general guard and public evaluation (\cref{sec:app_compare}).
\textbf{Review of the verifier boundary}: the scorer and the rule run from a frozen copy that agents cannot edit, the program rejects any code change that refers to development or test assets, the reviewer agent can veto every accepted change, and humans inspect the recipe of every champion.

\paragraph{Two stages.}
The loop ran in two stages.
Stage 1 uses injected defects, motion from optical flow and the semantics, quality and human-preference guards.
Stage 2 starts from the last stage-1 champion.
It adds the real dimension, the general guard, more subtle defect versions and family guards (\cref{sec:app_rsi}).
The champions carry over, and the rule keeps its form, with guards for the new families.
Humans designed this second exam after seeing where the first was too coarse; this curriculum step is the human part of the system.
Public-benchmark gains concentrate in this second stage (\cref{sec:exp_insight}).

%% file: fig/arch.tex
\begin{figure*}[t]
\centering
\includegraphics[width=\textwidth]{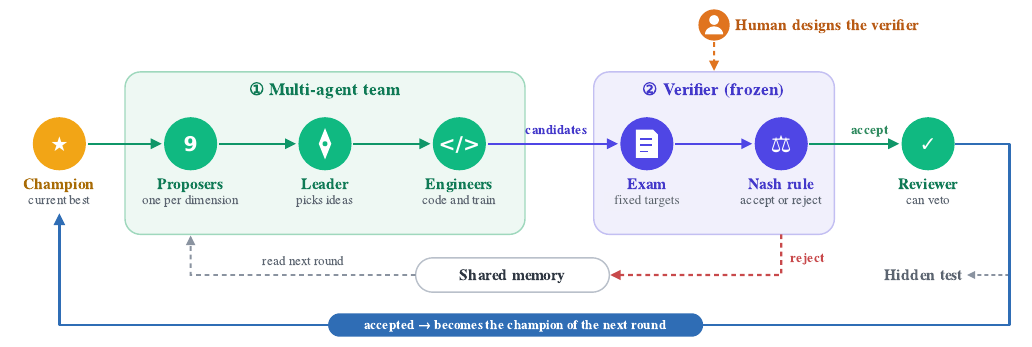}
\vspace{-5mm}
\caption{\textbf{The EVA-RSI system.}
\ding{172} A team of LLM agents proposes, implements and trains candidate evaluators from the current champion.
\ding{173} A frozen verifier scores each candidate on an exam with fixed targets, and a Nash bargaining rule accepts or rejects it.
An accepted candidate becomes the champion of the next round; every outcome enters a shared memory that the proposers read.
Humans design the verifier and redesign the exam between stages.}
\label{fig:arch}
\end{figure*}

%% file: sec/4_experiments.tex
\section{Evaluating the EVA-RSI design}
\label{sec:exp}

\subsection{Setup}
\label{sec:exp_setup}

We start from Qwen3-VL-8B-Instruct~\cite{qwen3vl} and run EVA-RSI in two stages, with 170 candidates over 88 rounds.
Stage 2 continues from the last stage-1 champion with the larger question bank (\cref{sec:method}).
Each candidate trains for at most 1.5 hours.
The agents run on Claude Opus 5.5, with fallbacks when it is unavailable (\cref{sec:app_rsi}).
The loop used 944 B200 GPU hours.
We score all eight champions on a hidden test bank of 3{,}450 questions.

\subsection{The exam guides evaluator improvement}
\label{sec:exp_rsi}

\input{fig/rsi_curve}
\input{fig/qual}

\begin{table}[t]
\centering
\small
\setlength{\tabcolsep}{4pt}
\resizebox{\linewidth}{!}{%
\begin{tabular}{lcccc}
\toprule
Dimension & Qwen3-VL-8B & After stage 1 & \textcolor{cOurs}{\textbf{\ours{}}} & Gain \\
\midrule
Physics      & 0.739 & \textbf{0.881} & 0.858 & \gain{18}{0.119} \\
Consistency  & 0.877 & 0.977 & \textbf{0.997} & \gain{18}{0.120} \\
Smoothness   & 0.461 & 0.841 & \textbf{0.888} & \gain{49}{0.427} \\
Dynamics     & 0.769 & 0.904 & \textbf{0.923} & \gain{21}{0.154} \\
Real videos  & 0.684 & 0.755 & \textbf{0.863} & \gain{24}{0.179} \\
\midrule
Semantics    & 0.988 & 0.992 & \textbf{0.992} & \gain{6}{0.004} \\
Quality      & 0.968 & 0.993 & \textbf{0.996} & \gain{9}{0.028} \\
Human pref.  & 0.624 & 0.640 & \textbf{0.660} & \gain{10}{0.036} \\
General      & 0.685 & 0.726 & \textbf{0.726} & \gain{10}{0.041} \\
\bottomrule
\end{tabular}}
\caption{\textbf{Normalized scores on the hidden test bank.}
Top: main dimensions. Bottom: guard dimensions.
Chance is $0.5$ for pairwise questions.}
\label{tab:rsi}
\end{table}

\Cref{fig:rsi_curve}a shows the test score of each champion.
The mean over the main dimensions rises from $0.706$ to $0.906$.

The largest gain is where Qwen3-VL-8B was weakest (\cref{tab:rsi}): smoothness rises from $0.461$ to $0.888$.
At the same 54-frame input length, Qwen3-VL-8B stays near chance.
Dynamics and physics also rise, so the evaluator judges both how much a video moves and whether its motion is plausible; physics peaks in stage 1 and gives back part of its gain in stage 2.
Scores also rise on real videos labeled by human raters.
The gains hold for single-video ratings, and pairwise answers depend less on video order.
No guard dimension drops from the starting model on the test bank.
\Cref{fig:qual} shows test pairs before and after RSI.

\textbf{The gains reach unseen defects.}
The test bank holds ten defect families of physics, consistency and smoothness that never appear in the development banks or the training data, such as teleporting objects, texture swaps and local frame shuffles.
On these families, the score rises from $0.754$ to $0.913$ (gain: $0.159$, 95\% interval $[0.116, 0.200]$), with the largest gain on smoothness.
The gain also holds without the questions that the stage-1 test bank shares (\cref{sec:app_unseen}).

\begin{table*}[t]
\centering
\footnotesize
\setlength{\tabcolsep}{4pt}
\begin{tabular}{llllp{0.52\textwidth}r}
\toprule
Stage & Champion & Role & Round & Recipe in one sentence & Test gain \\
\midrule
1 & c007 & dynamics & 5 & A weight average (``soup'') of three earlier candidates, which keeps their gains with less drift in human preference. & $+0.068$ \\
1 & c009 & physics & 5 & A weight soup plus a short LoRA polish with graded time-reversal labels and new physics and consistency defect families; proposed in the same round as c007 and judged against it after c007 was accepted. & $+0.142$ \\
1 & c022 & physics & 14 & A direction-aware vision tower: a time-antisymmetric change to the patch embedding, trained on motion and direction labels, language model frozen. & $+0.153$ \\
1 & c029 & semantics & 17 & Composes c022's vision tower with the language model of c021, at a higher resolution. & $+0.158$ \\
1 & c046 & human & 26 & The best 32-frame vision tower plus a language-model LoRA, with a repair step for human preference and semantics. & $+0.166$ \\
\midrule
2 & c001 & human & 1 & Fine-tuning on real generated videos with human labels, with base-model anchors for text alignment and replay of earlier defect families. & $+0.190$ \\
2 & c033 & general & 16 & A polish that adds position-debiased base-model answers on multiple-choice questions, which keeps general understanding. & $+0.196$ \\
2 & c071 (\ours{}) & physics & 30 & An arrow-of-time teacher (is this clip played forward or backward?), built from about 2{,}400 forward and reversed training clips at 54 frames and $230{,}400$ pixels, with a dynamics re-fit. & $+0.200$ \\
\bottomrule
\end{tabular}
\caption{\textbf{The eight champions found by RSI.} Role: the agent that proposed the candidate. Test gain: main mean minus the start model on the hidden test bank.}
\label{tab:lineage}
\end{table*}

\subsection{Dynamics on public benchmarks}
\label{sec:exp_compare}

\begin{table}[t]
\centering
\footnotesize
\setlength{\tabcolsep}{4pt}
\begin{tabular}{lcc}
\toprule
Model & VideoScore-Bench & VBench \\
\midrule
\oursrow \ours{} (ours) & \textbf{0.611} & \textbf{0.906} \\
After stage 1 (ours) & 0.564 & 0.781 \\
Qwen3-VL-8B (start) & 0.538 & 0.812 \\
\midrule
Qwen3-VL-32B & 0.495 & 0.875 \\
Qwen3.8-27B (thinking) & 0.453 & 0.750 \\
InternVL3.5-8B~\cite{wang2025internvl35} & 0.433 & 0.750 \\
VideoScore2~\cite{he2025videoscore2} & 0.534 & 0.406 \\
UnifiedReward-7B~\cite{wang2025unifiedreward} & 0.486 & 0.500 \\
VideoReward & $-$0.311 & 0.312 \\
\midrule
Claude Opus 5.5 & 0.549 & 0.781 \\
GPT-5.6 Sol & 0.575 & 0.844 \\
Muse Spark 1.3 & 0.512 & 0.844 \\
Grok 4.6 & 0.549 & 0.781 \\
Gemini 4 Argon & 0.480 & 0.719 \\
\bottomrule
\end{tabular}
\caption{\textbf{Video dynamics on public benchmarks.} Spearman correlation with human dynamics scores (VideoScore-Bench) and preference accuracy on dynamics pairs (VBench). 80 videos and 16 pairs from 6 prompts. Bold: highest observed score among 21 models (\cref{tab:compare}).}
\label{tab:dyn}
\end{table}

\Cref{tab:dyn} compares \ours{} with our stage-1 champion and 19 open models, trained video evaluators and proprietary models on the dynamics parts of two public benchmarks.
In VideoScore-Bench~\cite{he2024videoscore}, models rate the dynamic degree of generated videos, and we report the Spearman correlation with human ratings.
In VBench~\cite{huang2024vbench}, models pick the more dynamic of two generated videos, as human annotators did.
All models use their default inference settings (\cref{sec:app_compare}).

\textbf{\ours{} leads on the evaluated dynamics subsets.}
It has the highest point estimate among 21 models on both metrics (\cref{tab:dyn}).
All three stage-2 champions exceed the proprietary models on these two metrics.
\Cref{tab:compare} in \cref{sec:app_compare} gives the full comparison on 12 public benchmarks, where proprietary models lead on the overall mean and \ours{} leads all open models.
\Cref{fig:rsi_curve}c scores each champion on all 15 evaluation metrics of these benchmarks; the public score rises from $0.39$ to $0.71$, almost all in stage 2.

\label{sec:exp_insight}\textbf{Stage 2 improves public-benchmark performance.}
Without real-video questions, stage 1 raises the hidden-test score from $0.706$ to $0.872$, with the public score remaining near its starting level (\cref{fig:rsi_curve}a,c).
Stage 2 combines human-labeled real videos with subtler defects and new guards; its first champion raises the public score.

\subsection{Coordinating improvement across dimensions}
\label{sec:exp_agents}

\begin{table}[t]
\centering
\footnotesize
\setlength{\tabcolsep}{4pt}
\begin{tabular}{lrrr}
\toprule
Outcome & Stage 1 & Stage 2 & All \\
\midrule
Accepted & 5 & 3 & 8 \\
\midrule
Scored and rejected & 41 & 38 & 79 \\
\quad main gain too small & 31 & 32 & 63 \\
\quad dimension below floor & 10 & 5 & 15 \\
\quad a family guard & -- & 1 & 1 \\
\midrule
Not scored & 34 & 49 & 83 \\
\quad training job failed & 23 & 29 & 52 \\
\quad implementation rejected & 11 & 9 & 20 \\
\quad driver restart & 0 & 6 & 6 \\
\quad unfinished & 0 & 5 & 5 \\
\midrule
Total & 80 & 90 & 170 \\
\bottomrule
\end{tabular}
\caption{\textbf{Outcomes of the 170 candidates.}
Each rejected candidate counts under the first rule that it failed.}
\label{tab:outcomes}
\end{table}

Of the 170 candidates, 8 became champions (\cref{tab:lineage,tab:outcomes}).
The hidden-test point estimate rises at every change of champion; three of eight steps remain significant after Holm correction (\cref{sec:app_champ}).

In stage 1, the agents averaged the weights of earlier candidates, made the vision encoder sensitive to the direction of time, and added defect pairs in which both clips move at the same speed.
In stage 2, they fine-tuned on real generated videos, used the base model's answers as anchors, and trained on forward and reversed clips at the same 54-frame length used for evaluation.
\ours{} adds an arrow-of-time teacher to the previous champion (\cref{sec:app_evo}).
The dynamics score rises most when the agents add graded time-reversal labels in round 5.

\textbf{Improvement with skill retention.}
One LoRA trained once on all the loop's training data, including the stage-2 pools (Union SFT), reaches $0.889$ on test and $0.558$ on human preference, versus $0.906$ and $0.660$ for \ours{}; its public score is $0.53$ against $0.71$.
Adding anchors satisfies the rule but reaches $0.848$ and $0.844$ across two seeds (\cref{sec:app_once}). These controls expose the trade-off addressed by coordinated search: fitting the pooled data improves motion scores, while retaining other skills requires further recipe choices.

In a non-recursive control, the same agents trained each candidate from Qwen3-VL-8B using the same data and acceptance rule. They saw earlier proposals' titles and roles, but no scores or outcomes.
One of ten proposals failed a budget check; the other nine missed human-preference or general-understanding floors. \ours{} outperformed the development-selected candidate and all four tested candidates on the hidden test (\cref{tab:app_once}).

\paragraph{The final recipe uses inherited capabilities.}
Replaying the c033 recipe directly from Qwen3-VL-8B yields test means of $0.757$ and $0.756$ across two seeds; c033 trained along the champion chain reaches $0.902$ (\cref{tab:app_once,tab:champions}).
The difference is pronounced on smoothness: $0.514$ and $0.505$ from direct training versus $0.850$ along the chain.
The recipe thus uses accumulated vision and motion capabilities. Compute differs: 944 GPU hours for EVA-RSI, 384 for Union SFT, 352 per Union + anchors seed and 212 for the non-recursive search, stopped after observed stagnation. The controls assess the complete search design at these budgets.

\subsection{Lessons from running RSI}
\label{sec:exp_lessons}

\textbf{Exam coverage shapes the gains.}
Real-video scores stayed between $0.700$ and $0.755$ through stage 1; the first champion under the expanded stage-2 exam raised the score to $0.852$.
\textbf{Returns diminish on a fixed exam.}
All stage-2 candidates scored after \ours{} landed within $0.004$ of it, motivating a broader exam for future search (more in \cref{sec:app_lessons}).

%% file: fig/rsi_curve.tex
\begin{figure*}[t]
\centering
\includegraphics[width=\textwidth]{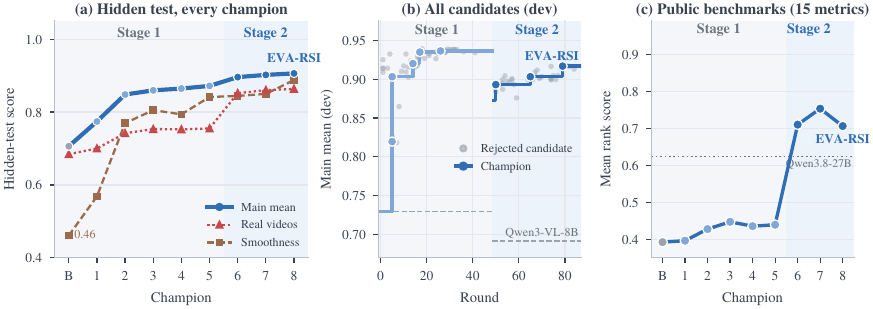}
\vspace{-6mm}
\caption{\textbf{RSI improves the evaluator over rounds.}
(a) Normalized score of each champion on the hidden test bank.
(b) Scored candidates (grey) and champions (blue) on the development banks, which change at the start of stage 2.
(c) Mean rank score on 12 public benchmarks, against all evaluators except VideoReward (\cref{sec:app_compare}). The development banks determine the final champion.}
\label{fig:rsi_curve}
\end{figure*}

%% file: fig/qual.tex
\begin{figure*}[t]
\centering
\includegraphics[width=\textwidth]{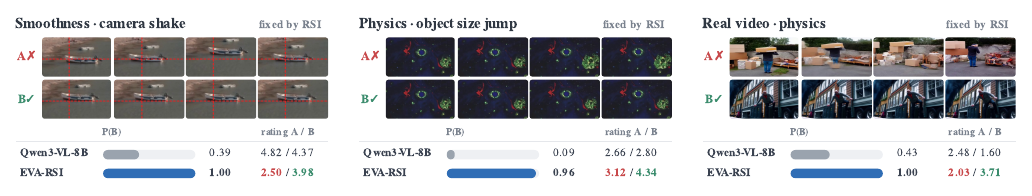}
\vspace{-6mm}
\caption{\textbf{Test-bank pairs before and after RSI.}
B is the correct answer in each pair.
P is the probability that each model gives to B.
Left: close-up of four consecutive frames; dashed lines mark a fixed position.
Middle: objects suddenly change size.
Right: people and boxes vanish and reappear.}
\label{fig:qual}
\end{figure*}

%% file: sec/5_downstream.tex
\section{\ours{} as a reward and a data filter}
\label{sec:downstream}

We use \ours{} as an RL reward and as a filter for training data. Both pipelines increase motion in generated videos; human raters prefer the RL results at update 50.

\begin{table}[t]
\centering
\footnotesize
\setlength{\tabcolsep}{2.2pt}
\resizebox{\linewidth}{!}{%
\begin{tabular}{llccccc}
\toprule
Reward & Upd. & Dyn.\,$\uparrow$ & Flow\,$\uparrow$ & Obj.\,$\uparrow$ & Adh.\,$\uparrow$ & Elem.\,$\uparrow$ \\
\midrule
None (start) & 0 & 0.579 & 13.77 & 1.43 & 3.668 & 0.745 \\
\midrule
Qwen3-VL-8B, seed 1 & 50 & 0.790 & 15.64 & 1.67 & 3.741 & 0.762 \\
Qwen3-VL-8B, seed 2 & 50 & 0.749 & 14.97 & 1.49 & 3.684 & 0.751 \\
\oursrow \ours{}, seed 1 & 50 & 0.848 & 17.64 & 1.76 & 3.730 & 0.761 \\
\oursrow \ours{}, seed 2 & 50 & \textbf{0.892} & \textbf{19.81} & \textbf{2.03} & 3.714 & 0.757 \\
\midrule
Qwen3-VL-8B, seed 1 & 100 & 0.770 & 14.93 & 1.37 & 3.754 & 0.761 \\
Qwen3-VL-8B, seed 2 & 100 & 0.700 & 13.48 & 1.34 & 3.671 & 0.748 \\
\oursrow \ours{}, seed 1 & 100 & \textbf{0.913} & \textbf{20.53} & \textbf{1.88} & 3.675 & 0.750 \\
\oursrow \ours{}, seed 2 & 100 & 0.892 & 17.59 & 1.62 & 3.519 & 0.712 \\
\midrule
VideoReward (fair), seed 1 & 50 & 0.351 & 7.94 & 0.82 & 3.041 & 0.617 \\
VideoReward (fair), seed 2 & 50 & 0.271 & 7.33 & 0.69 & 2.956 & 0.604 \\
VideoReward & 50 & 0.289 & 6.73 & 0.70 & 2.575 & 0.522 \\
VideoReward & 120 & 0.095 & 4.06 & 0.52 & 1.690 & 0.307 \\
\midrule
\multicolumn{7}{l}{\emph{Trained to the full budget (500 GPU hours)}} \\
Qwen3-VL-8B & 346 & 0.778 & 21.08 & 2.00 & 1.841 & 0.265 \\
\oursrow \ours{} & 342 & \textbf{0.987} & \textbf{26.34} & \textbf{2.79} & 3.114 & 0.616 \\
\bottomrule
\end{tabular}}
\caption{\textbf{RL with different rewards} (630 test videos).
Dyn.: dynamic degree. Obj.: object motion. Adh.\ and Elem.: judge scores.
VideoReward (fair): same penalties and settings as \ours{}. VideoReward: stress test without penalties.}
\label{tab:main}
\end{table}

\input{fig/ds_qual}

\begin{table}[t]
\centering
\footnotesize
\setlength{\tabcolsep}{4pt}
\begin{tabular}{lcc}
\toprule
Question & vs.\ Qwen3-VL-8B & vs.\ VideoReward (fair) \\
\midrule
More motion & 0.77 $[0.68, 0.85]$ & 0.85 $[0.75, 0.94]$ \\
Better overall & 0.73 $[0.65, 0.80]$ & 0.69 $[0.55, 0.82]$ \\
\midrule
Visual quality & 0.71 $[0.63, 0.79]$ & 0.66 $[0.51, 0.79]$ \\
Prompt adherence & 0.73 $[0.65, 0.80]$ & 0.71 $[0.58, 0.82]$ \\
\bottomrule
\end{tabular}
\caption{\textbf{Human study at update 50 (training seed 1)}, share of answers for \ours{} (0.5: no preference). Six raters; 48 and 24 pairs.}
\label{tab:app_human}
\end{table}

\subsection{RL reward}
\label{sec:setup}

All runs fine-tune Wan2.1 T2V-1.3B~\cite{wan2025} with DiffusionART~\cite{diffusionart}, a group-based policy-gradient method; all runs share the settings that we chose in pilots with \ours{}, and only the reward model changes.
The \ours{} reward sums its 1--5 ratings for dynamics, physics, smoothness and quality, with fixed penalties for flicker and uneven global motion (\cref{sec:app_rl}).
We compare it with the evaluator before RSI, Qwen3-VL-8B, which reads the same 54 frames, so the two rewards differ only in their weights.
We also run VideoReward~\cite{liu2025videoreward} with the same penalties and settings (fair) and without penalties (stress test).
We measure motion with VBench~\cite{huang2024vbench} \emph{dynamic degree} (RAFT~\cite{teed2020raft}), \emph{optical flow}~\cite{farneback2003flow} and \emph{object motion}, the flow that remains after we remove camera motion.
A blind Qwen3.8-27B judge rates prompt \emph{adherence}.
We test on 630 videos from VBench and VidProM~\cite{wang2024vidprom} prompts, and confirm each result on 300 MovieGenBench~\cite{polyak2024moviegen} prompts.
\label{sec:exp_main}

\paragraph{More motion.}
All 32 motion tests pass after Holm correction: for both seeds, updates 50 and 100 and both prompt sets, \ours{} gives more motion than the start model and than Qwen3-VL-8B (\cref{tab:main}, \cref{sec:app_rl_stats}).
Object motion rises in every comparison after camera motion is removed.
The gain tracks the loop: the stage-1 champion increases motion over Qwen3-VL-8B at update 50; \ours{} sustains a significant lead at update 100 (\cref{sec:app_rl}).
At update 50, adherence matches Qwen3-VL-8B; it falls with longer training (\cref{sec:app_rl}).

\paragraph{Better motion.}
Six raters compared pairs of videos from training seed 1 at update 50 with the same prompt and sampling seed. We randomized their left--right order and counted ties as half a vote.
Raters judged \ours{} videos to move more than Qwen3-VL-8B videos and to be better overall, in visual quality and in adherence (\cref{tab:app_human}).
When they see more motion in the \ours{} video, they also prefer it overall in nine of ten answers.

\paragraph{Motion quality beyond flow magnitude.}
A raw optical-flow reward with the same settings moves more than \ours{} at update 50, but its videos jitter more, and by update 100 \ours{} leads it in both motion and adherence (\cref{sec:app_rl}).

\paragraph{VideoReward reduces motion under these settings.}
VideoReward, which is anti-correlated with human dynamics ratings (\cref{tab:dyn}), does the opposite.
Its matched-setting run reduces dynamic degree to $0.351$ and $0.271$ from the start model's $0.579$ (\cref{tab:main}), and raters prefer \ours{} over it on motion and overall.
Without penalties, it collapses to still close-ups of faces (\cref{fig:overview,fig:ds_qual}).

\subsection{Data filter}
\label{sec:exp_sft}

\begin{table}[t]
\centering
\footnotesize
\setlength{\tabcolsep}{4pt}
\begin{tabular}{lcc}
\toprule
Picked $-$ random & Step 1{,}500 & Step 3{,}000 \\
\midrule
Dynamic degree\,$\uparrow$ & $+0.143^*$ & $+0.277^*$ \\
Optical flow\,$\uparrow$ & $+3.47^*$ & $+7.38^*$ \\
Adherence\,$\uparrow$ & $+0.10$ & $+0.03$ \\
Elements\,$\uparrow$ & $+0.022$ & $-0.001$ \\
\midrule
Dyn.\ degree, picked / random & 0.867 / 0.723 & 0.873 / 0.597 \\
\bottomrule
\end{tabular}
\caption{\textbf{Selection pipeline incorporating \ours{} versus random data} (300 prompts). $^*$: 95\% interval excludes zero.}
\label{tab:sft}
\end{table}

We fine-tune Wan2.1 on 50k clips picked by a pipeline that ranks clips with \ours{} after flow and caption filters, and on 50k random clips from a different pool, with the same recipe to evaluate the selection pipeline (\cref{sec:app_sft}).
The flow threshold and caption filter select moving clips without scene cuts. The two training sets differ in source composition and share about one fifth of their clips. We evaluate both models on the same 300 prompts.
After 3{,}000 steps, the model trained on selected data produces motion in $87\%$ of its videos, compared with $60\%$ for the model trained on random data, with an adherence difference of $+0.03$ whose 95\% interval includes zero (\cref{tab:sft}).

\subsection{Limitations and future work}
\label{sec:discussion}

The main limitation of EVA-RSI is its verifier exam. Controlled edits simplify motion failures and differ from the distribution of real generated videos, where defects interact with scene content and intent. Construction-defined targets describe individual edits. Human-labeled real videos broaden exam coverage; complex, interacting failures remain a target for expansion. Future work should build richer exams from generator failures, validate targets with human judgments, and test transfer across generators.

%% file: fig/ds_qual.tex
\begin{figure*}[t]
\centering
\includegraphics[width=\textwidth]{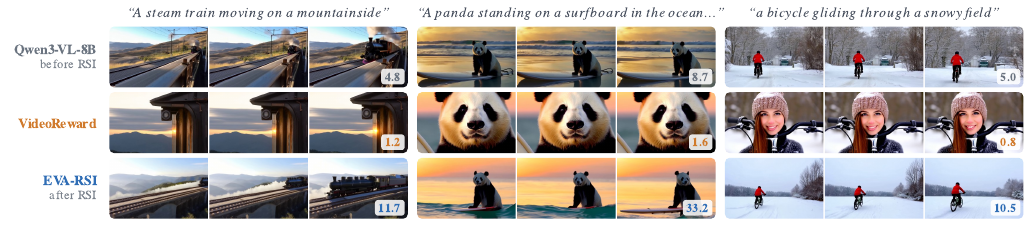}
\vspace{-5mm}
\caption{\textbf{Same prompt and seed; only the reward changes} (update 50).
\ours{} keeps the scene and gives the most motion. VideoReward: visual and motion quality, without penalties. Corner: optical flow.}
\label{fig:ds_qual}
\end{figure*}

%% file: sec/5_conclusion.tex
\section{Conclusion}
\label{sec:conclusion}

EVA-RSI explores how recursive improvement can extend to video evaluation through two designs: a constructed-target exam supplies feedback, and dimension-specific agents with Nash-based acceptance coordinate competing objectives. The resulting search improves temporal judgments while retaining guarded skills, transfers to unseen defects, and produces an evaluator that benefits video generation through RL rewards and data selection. This study establishes the practical promise of evaluator self-improvement and identifies richer, human-validated exams as a central direction for further progress.

\label{body_end}

%% file: appendix.tex
\begin{center}\Large\bfseries Appendix\end{center}
\appendix
\crefalias{section}{appendix}
\crefalias{subsection}{subappendix}
\setcounter{figure}{0}
\setcounter{table}{0}
\setcounter{equation}{0}
\renewcommand{\thefigure}{S\arabic{figure}}
\renewcommand{\thetable}{S\arabic{table}}
\renewcommand{\theequation}{S\arabic{equation}}
\renewcommand{\theHfigure}{appendix.\arabic{figure}}
\renewcommand{\theHtable}{appendix.\arabic{table}}
\renewcommand{\theHequation}{appendix.\arabic{equation}}
\input{sec/X_appendix}

%% file: sec/X_appendix.tex

\section{RSI details}
\label{sec:app_rsi}

\subsection{The evaluator}
The evaluator is a vision-language model that receives one or two videos and a question and answers with one token: \texttt{A} or \texttt{B} for a pairwise question, a rating from \texttt{1} to \texttt{5}, or an option letter.
We read each answer from the next-token probabilities, so a rating is the expected value over \texttt{1}\dots\texttt{5}.
The loop can change its weights, in practice with LoRA fine-tuning, and its inference settings, such as the system prompt and the number and size of frames.

\subsection{The two stages}

EVA-RSI ran in two stages (\cref{tab:rsi_runs}).
Stage 2 starts from the last stage-1 champion and uses a larger question bank.
Its last champion is \ours{}, which we use in all downstream experiments.

\begin{table}[ht]
\centering
\footnotesize
\setlength{\tabcolsep}{4pt}
\begin{tabular}{lcccc}
\toprule
Stage & Start & Rounds & Candidates & Champions \\
\midrule
1 & Qwen3-VL-8B & 49 & 80  & 5 \\
2 & stage-1 champion & 39 & 90  & 3 \\
\midrule
Total & & 88 & 170 & 8 \\
\bottomrule
\end{tabular}
\caption{\textbf{The two stages of EVA-RSI.}}
\label{tab:rsi_runs}
\end{table}

\paragraph{Stage 1.}
Stage 1 started from Qwen3-VL-8B-Instruct and ran for about 39 hours on 16 B200 GPUs.
Stage 1 raised real-video test score from $0.684$ to $0.755$.
The base model already scored near ceiling on several development families.

\paragraph{Stage 2.}
Stage 2 added the real-video and general dimensions, subtler defects, harder motion and semantic pairs, and guards for single families (\cref{sec:app_rule}).
It ran for about 47 hours on 16 and then 32 B200 GPUs.
\ours{} reads 54 frames per video.
Its last recipe teaches it to tell forward from reversed clips, while keeping the other dimensions close to the previous champion.

\paragraph{A third stage.}
We also ran a third stage from \ours{} with the same verifier and a larger test bank.
None of its 100 candidates met the rule.
Midway, we added a looser rule, and two candidates passed it; the better one is level with \ours{} on the hidden test bank, so we keep \ours{}.
When the questions are nearly solved, more rounds on them no longer help.

\paragraph{Step sizes.}
\label{sec:app_champ}
On the stage-2 hidden test bank, the main mean rises at every change of champion.
\Cref{tab:app_steps} tests each of the eight steps with a paired bootstrap.
Three steps stay significant after a Holm correction: the first two champions (c007 and c009) and the first stage-2 champion (c001).
Physics rises significantly at the first step ($+0.118$, interval $[+0.085, +0.151]$).
The step from c022 to c029 also raises the resolution from 147k to 230k pixels per frame.
Stage-1 champions did not see the real-video and general dimensions during acceptance.

\begin{table}[ht]
\centering
\footnotesize
\setlength{\tabcolsep}{4pt}
\begin{tabular}{lccc}
\toprule
Step & Gain in main mean & 95\% interval & Holm \\
\midrule
Start $\to$ c007 & $+0.068$ & $[+0.054, +0.082]$ & \checkmark \\
c007 $\to$ c009 & $+0.074$ & $[+0.056, +0.091]$ & \checkmark \\
c009 $\to$ c022 & $+0.011$ & $[+0.001, +0.022]$ & \\
c022 $\to$ c029 & $+0.005$ & $[-0.008, +0.019]$ & \\
c029 $\to$ c046 & $+0.007$ & $[-0.002, +0.017]$ & \\
c046 $\to$ c001 & $+0.024$ & $[+0.012, +0.037]$ & \checkmark \\
c001 $\to$ c033 & $+0.006$ & $[-0.002, +0.014]$ & \\
c033 $\to$ c071 & $+0.004$ & $[-0.008, +0.016]$ & \\
\bottomrule
\end{tabular}
\caption{\textbf{Steps between champions} (stage-2 hidden test bank). Holm: significant after correction over the eight steps.}
\label{tab:app_steps}
\end{table}

\begin{table*}[t]
\centering
\footnotesize
\setlength{\tabcolsep}{4pt}
\begin{tabular}{l c ccccc ccc}
\toprule
& & \multicolumn{5}{c}{Stage 1} & \multicolumn{3}{c}{Stage 2} \\
\cmidrule(lr){3-7}\cmidrule(lr){8-10}
Dimension & Base & c007 & c009 & c022 & c029 & c046 & c001 & c033 & c071 (\ours{}) \\
\midrule
Proposed in round & -- & 5 & 5 & 14 & 17 & 26 & 1 & 16 & 30 \\
Frames & 32$^\dagger$ & 32 & 32 & 32 & 32 & 32 & 32 & 32 & 54 \\
Pixels per frame (k) & 147$^\dagger$ & 147 & 147 & 147 & 230 & 230 & 230 & 230 & 230 \\
\midrule
Physics      & 0.739 & 0.857 & 0.874 & 0.868 & 0.893 & 0.881 & 0.886 & 0.868 & 0.858 \\
Consistency  & 0.877 & 0.943 & 0.960 & 0.968 & 0.977 & 0.977 & 0.995 & 0.998 & 0.997 \\
Smoothness   & 0.461 & 0.569 & 0.770 & 0.805 & 0.794 & 0.841 & 0.844 & 0.850 & 0.888 \\
Dynamics     & 0.769 & 0.802 & 0.896 & 0.903 & 0.906 & 0.904 & 0.902 & 0.934 & 0.923 \\
Real videos  & 0.684 & 0.700 & 0.742 & 0.753 & 0.752 & 0.755 & 0.852 & 0.860 & 0.863 \\
\midrule
Semantics    & 0.988 & 0.988 & 0.992 & 0.996 & 0.996 & 0.992 & 0.988 & 0.988 & 0.992 \\
Quality      & 0.968 & 0.977 & 0.994 & 0.995 & 0.992 & 0.993 & 0.994 & 0.997 & 0.996 \\
Human pref.  & 0.624 & 0.622 & 0.621 & 0.623 & 0.631 & 0.640 & 0.656 & 0.656 & 0.660 \\
General      & 0.685 & 0.704 & 0.723 & 0.730 & 0.715 & 0.726 & 0.711 & 0.719 & 0.726 \\
\midrule
Main mean    & 0.706 & 0.774 & 0.848 & 0.860 & 0.865 & 0.872 & 0.896 & 0.902 & 0.906 \\
\bottomrule
\end{tabular}
\caption{\textbf{All eight champions on the hidden test bank.} Main: mean of the five main dimensions. $^\dagger$Default settings.}
\label{tab:champions}
\end{table*}

\paragraph{Test bank.}
The stage-2 prompts name one weak point seen on the stage-1 test bank, reversal of slow clips (without these 36 test questions the final main mean is $0.916$, not $0.906$), and the two stages' test banks share $1{,}254$ questions outside the real and general dimensions.

\subsection{Full acceptance rule}
\label{sec:app_rule}

Each dimension $i$ has utility $u_i = s_i - d_i$.
Here $d_i$ is the score of the untouched Qwen3-VL-8B-Instruct minus a tolerance of $0.02$ ($0.015$ for human preference).
A candidate replaces the champion only if all of the following hold:
\begin{enumerate}
\item every dimension stays above its floor, $u_i > 0$;
\item the mean of the main dimensions rises by at least $0.01$, with a 90\% lower bound above 0 under a paired cluster bootstrap;
\item no main dimension drops by more than $0.03$;
\item the weighted log Nash product rises (\cref{eq:nash}).
\end{enumerate}
The main dimensions are physics, consistency, smoothness, dynamics and real videos, with weight 1.
Semantics, quality and general have weight $0.5$, and human preference has weight $0.75$.
A second acceptance rule applies when the main dimensions saturate.
It accepts a candidate whose main scores stay within $0.005$ of the champion and whose weighted gain on the other dimensions is at least $0.005$, with a bootstrap lower bound above 0.
Rules 1 and 4 still apply.
The last stage-1 champion (c046) was accepted through this path.
In stage 2, no family with at least 40 questions may drop by more than $0.12$, and no VideoGen-RewardBench aspect by more than $0.05$.
A reviewer agent can still veto an accepted candidate.

\paragraph{Scoring.}
A pairwise question scores 1, 0.5 (tie) or 0.
A rating question with target $t$ scores $1 - |\mathbb{E}[r] - t|/4$, and a multiple-choice question scores 1 or 0.
A dimension score is the mean over its families of the mean question score.
Motion pairs use clips whose optical flow differs by at least $2.5\times$ ($1.5$--$2.5\times$ for hard pairs).
Real-video pairs use generated videos whose human scores differ by at least 2 points.

\subsection{Unseen defects and shared questions}
\label{sec:app_unseen}

Ten test families of the main dimensions never appear in the development banks or the training data: crossfade drift (two strengths) and object texture swap for consistency; object scale pop, teleport (two sizes) and upside-down clips for physics; and local frame shuffles (two densities) and speed bursts for smoothness.
\Cref{tab:app_unseen} shows that \ours{} gains $0.159$ on these 440 questions (95\% interval $[0.116, 0.200]$), compared with $0.247$ on seen families. Smoothness has the largest unseen-family gain.
The two stages' test banks share 1{,}254 questions; without them, the main-mean gain is $0.194$, compared with $0.200$ on the full bank.

\begin{table}[ht]
\centering
\footnotesize
\setlength{\tabcolsep}{4pt}
\begin{tabular}{lccl}
\toprule
Questions & Start & \ours{} & Gain [95\% int.] \\
\midrule
Seen families & 0.662 & 0.909 & $+0.247$ \\
Unseen families & 0.754 & 0.913 & $+0.159$ [0.116, 0.200] \\
\quad physics & 0.849 & 0.943 & $+0.094$ \\
\quad consistency & 0.985 & 1.000 & $+0.015$ \\
\quad smoothness & 0.428 & 0.795 & $+0.367$ [0.252, 0.479] \\
\midrule
Main mean, all & 0.706 & 0.906 & $+0.200$ [0.181, 0.219] \\
Main mean, not shared & 0.709 & 0.902 & $+0.194$ [0.172, 0.216] \\
\bottomrule
\end{tabular}
\caption{\textbf{Unseen defects and shared questions} on the hidden test bank. Families: mean over physics, consistency and smoothness. Not shared: without the questions that the stage-1 test bank also contains. Intervals: bootstrap over clip clusters.}
\label{tab:app_unseen}
\end{table}

\subsection{Replaying the acceptance rule on the hidden test}
\label{sec:app_replay}

During the loop, only champions were scored on the hidden test bank.
Afterwards, we also scored the stage-2 candidates that were rejected despite a clear development gain.
These candidates failed only on the Nash product or a family drop.
\Cref{tab:replay} compares each of them with its parent.
Their test means stay close to their parents' scores. All three candidates rejected by the Nash criterion lose score on physics.

\paragraph{Simpler acceptance rules.}
For these recorded candidates, removing the Nash condition changes three decisions to acceptance. Their test-mean changes range from $-0.002$ to $+0.003$, with lower physics scores in each case. This replay diagnoses decisions on a fixed trajectory; a new search under another rule would generate a different trajectory.
A rule of mean gain alone would also have accepted the best of the first ten stage-1 candidates (c006), which does not transfer to public benchmarks (\cref{tab:app_once}).

\begin{table}[t]
\centering
\scriptsize
\setlength{\tabcolsep}{2.5pt}
\resizebox{\linewidth}{!}{%
\begin{tabular}{lllcccc}
\toprule
Cand. & Parent & Decision & Dev gain & Test main & $\Delta$ main & $\Delta$ human \\
\midrule
c010 & c001 & Nash & $+0.011$ & 0.896 & $+0.000$ & $-0.007$ \\
c035 & c001 & Nash & $+0.011$ & 0.894 & $-0.002$ & $-0.003$ \\
c056 & c033 & Nash & $+0.013$ & 0.905 & $+0.003$ & $-0.006$ \\
c033 (54 frames) & c033 & family drop & $+0.011$ & 0.904 & $+0.002$ & $+0.002$ \\
\midrule
\textbf{c001} & stage-1 champ. & accepted & $+0.021$ & 0.896 & $+0.024$ & $+0.016$ \\
\textbf{c033} & c001 & accepted & $+0.010$ & 0.902 & $+0.006$ & $+0.000$ \\
\textbf{c071} & c033 & accepted & $+0.014$ & 0.906 & $+0.004$ & $+0.004$ \\
\bottomrule
\end{tabular}}
\caption{\textbf{Acceptance replay on the stage-2 hidden test bank.} Top: rejected candidates with a development gain of at least $0.01$. Bottom: champions. $\Delta$: test score minus the parent.}
\label{tab:replay}
\end{table}

\subsection{Baselines without recursion}
\label{sec:app_once}

\begin{table*}[t]
\centering
\scriptsize
\setlength{\tabcolsep}{2.2pt}
\resizebox{\textwidth}{!}{%
\begin{tabular}{lcccccccccccccl}
\toprule
Model & Frames & Phys. & Cons. & Smooth. & Dyn. & Real & Sem. & Qual. & Human & Gen. & Main & Public & GPU h & Rule \\
\midrule
Qwen3-VL-8B & 32 & 0.739 & 0.877 & 0.461 & 0.769 & 0.684 & 0.988 & 0.968 & 0.624 & 0.685 & 0.706 & 0.39 & -- & -- \\
Union SFT & 54 & 0.859 & 0.977 & 0.844 & 0.904 & 0.859 & 0.992 & 0.984 & 0.558 & \textbf{0.726} & 0.889 & 0.53 & 384 & human floor \\
Real-only SFT & 54 & 0.844 & 0.907 & 0.608 & 0.751 & 0.862 & 0.992 & 0.960 & 0.598 & 0.715 & 0.794 & 0.57 & 6.2 & human floor \\
Union + anchors, seed 1 & 54 & 0.867 & 0.980 & 0.683 & 0.853 & 0.856 & 0.992 & 0.978 & 0.647 & 0.715 & 0.848 & 0.66 & 352 & accept \\
Union + anchors, seed 2 & 54 & \textbf{0.873} & 0.969 & 0.679 & 0.847 & 0.854 & 0.992 & 0.964 & 0.654 & 0.723 & 0.844 & 0.60 & 352 & accept \\
c033 recipe, seed 1 & 32 & 0.850 & 0.915 & 0.514 & 0.772 & 0.733 & 0.992 & 0.965 & 0.632 & 0.704 & 0.757 & 0.46 & 7.2 & general floor \\
c033 recipe, seed 2 & 32 & 0.851 & 0.915 & 0.505 & 0.775 & 0.737 & 0.992 & 0.969 & 0.608 & 0.700 & 0.756 & 0.45 & 7.3 & accept \\
Best of 10 (c006) & 32 & 0.845 & 0.957 & 0.854 & 0.910 & 0.717 & \textbf{0.996} & 0.995 & 0.611 & 0.704 & 0.857 & 0.40 & $\le$60 & human floor \\
\midrule
Non-recursive search (c004) & 32 & 0.842 & 0.975 & 0.726 & 0.916 & 0.862 & \textbf{0.996} & 0.995 & 0.610 & 0.704 & 0.864 & -- & 212 & human floor \\
\midrule
\oursrow \ours{} & 54 & 0.858 & \textbf{0.997} & \textbf{0.888} & \textbf{0.923} & \textbf{0.863} & 0.992 & \textbf{0.996} & \textbf{0.660} & \textbf{0.726} & \textbf{0.906} & \textbf{0.71} & 944 & accepted \\
\bottomrule
\end{tabular}}
\caption{\textbf{Baselines without the loop on the hidden test bank.}
Main: mean of the five main dimensions. Public: mean rank score as in \cref{fig:rsi_curve}c. Rule: decision on the development banks (\cref{sec:app_once}). Bold: best.}
\label{tab:app_once}
\end{table*}

\Cref{tab:app_once} lists every model that we trained from Qwen3-VL-8B without the loop.
Each model is scored on the stage-2 test bank with its own frame count.
\begin{itemize}
\item \textbf{Union SFT} and \textbf{real-only SFT} train one LoRA on all the loop's training data, or on its real videos only.
\item \textbf{Union + anchors} adds the anchor answers that c033 uses to keep human preference and general understanding.
\item \textbf{c033 recipe} repeats the full recipe of the c033 step, but starts from Qwen3-VL-8B and not from the champion before it.
\item \textbf{Best of 10} is the best, by development gain, of the first ten stage-1 candidates (c001--c010, rounds 1--6); candidate IDs follow proposal order.
\end{itemize}
The last column applies the acceptance rule with Qwen3-VL-8B as the champion.
Three of these models would be accepted.
Union SFT used 384 GPU hours. Each Union + anchors seed used 352 GPU hours on 32 GPUs (11 wall-clock hours). The full EVA-RSI search used 944 GPU hours; these comparisons evaluate resulting models at different compute budgets.
Union SFT uses all training pools, including the stage-2 pools that humans built after stage 1.
It reaches $0.889$ on test, fails the human-preference floor and transfers less to public benchmarks; \ours{} is significantly better on test and on human preference (95\% intervals exclude zero).
Real-only SFT reaches $0.57$ on the public score and $0.794$ on test.
Union + anchors passes the rule and trails \ours{} by $0.058$ and $0.062$ on test across two seeds, with the largest gap on smoothness.

\paragraph{Non-recursive search.}
The non-recursive control used the same 16 GPUs and development banks, judging candidates against Qwen3-VL-8B. Its context omitted leader notes, scores and outcomes. We stopped after observed stagnation at 212 GPU hours, short of the full loop's 944 GPU hours.
Since no candidate passes, we score on the test bank the three best by development main mean and the one closest to passing (c008).
\ours{} is significantly above both the pick by development gain (c004, \cref{tab:app_once}) and the best of the four on test (c005).

\section{How the evaluator evolved}
\label{sec:app_evo}

During the search, test evaluation covered champions. Post-search analyses also evaluated rejected candidates and controls (\cref{sec:app_replay,sec:app_once}).

\subsection{Champion lineage}

\Cref{tab:lineage} in the main paper lists the eight champions with their recipes.
Most champions build on the previous one, and several combine parts of earlier candidates.

\subsection{How the candidates ended}

\begin{figure*}[t]
\centering
\includegraphics[width=\textwidth]{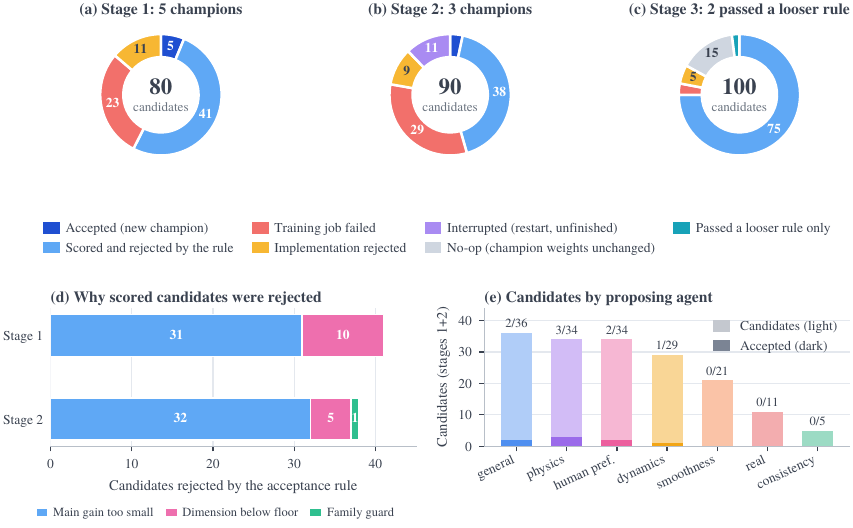}
\vspace{-3mm}
\caption{\textbf{Outcomes of the candidates.} (a--c) Outcome per stage. In stage 3 (\cref{sec:app_rsi}), no candidate met the frozen rule, two passed only a looser rule (teal), and three training jobs failed (unlabelled slices). (d) Why scored candidates were rejected, stages 1 and 2. (e) Candidates by proposing agent, stages 1 and 2; the dark part of each bar counts the accepted ones; the semantics and quality roles of stage 1 are counted as general.}
\label{fig:rsi_stats}
\end{figure*}

\Cref{tab:outcomes} and \cref{fig:rsi_stats}(a, b) sort the 170 candidates of stages 1 and 2 by how they ended.
Eight became champions.
Most scored candidates were rejected because their gain was too small.
About half of all candidates never reached scoring, mostly because their training job failed or their code was rejected.
\Cref{fig:rsi_stats}(c) adds the 100 candidates of stage 3.

\paragraph{Roles.}
The agents did not propose equally often.
In stage 1, the human-preference agent proposed about a third of the candidates.
In stage 2, the general and physics agents proposed the most.

\subsection{What worked}

\paragraph{Smoothness in stage 1.}
The second champion (c009) gave the largest single gain in smoothness.
It added graded time-reversal labels and defect pairs in which both clips move at the same speed.
Matching the speed forces the model to distinguish the clips by their motion defects.

\paragraph{Real generated videos in stage 2.}
The first stage-2 champion (c001) trained on real generated videos with human labels.
It gave the largest gain on real videos, on both the development and the test bank.

\paragraph{Keeping general understanding.}
In stage 2, general understanding was the dimension closest to its floor.
c033 added the base model's own answers to multiple-choice questions, with shuffled answer order, as anchors.
This kept general understanding above its floor while the main dimensions still rose.

\paragraph{Telling forward from backward.}
\ours{} (c071) learned from a teacher whether a clip plays forward or backward.
The teacher is built from forward and reversed clips of the training split and sees the same 54 frames that \ours{} is scored on.
This step raised smoothness from $0.850$ to $0.888$ and the main mean from $0.902$ to $0.906$.

\subsection{What failed}

\paragraph{Trading one family for the mean.}
In stage 1, the rule saw only the means of the dimensions.
So it accepted c009 although one family (time reversal) fell sharply, because the main mean rose.
Later the same family fell again on the test bank.
Stage 2 therefore added guards for single families (\cref{sec:app_rule}).

\paragraph{Physics drifting in stage 2.}
Across the last two champions, test physics changed from $0.886$ to $0.858$ and the main mean from $0.896$ to $0.906$ (\cref{tab:champions}).

\paragraph{Floors and failed jobs.}
The human-preference floor rejected many candidates, mostly in stage 1.
Many training jobs failed, and some implementations were rejected before training (\cref{tab:outcomes}).

\paragraph{Text alignment.}
No candidate that targeted text alignment was accepted.
This matches the public benchmarks: on text alignment, \ours{} scores above its starting model on AIGVE and below it on VideoScore-Bench-v2 and below VideoScore2 (\cref{tab:compare}).

\subsection{Agents and cost}

The agents are headless coding-agent sessions, one per role and call.
Each has a fixed role prompt, can read the champion's development scores and past candidates, and can write only the candidate's code.
Stage 1 used Claude Opus 5.5, with GPT-5.3-Codex as a fallback on usage limits.
Stage 2 used Claude Opus 5.5, with Claude Sonnet 5.5 and Claude Opus 4.6 as fallbacks.
Training and scoring of candidates used about 940 B200 GPU hours (358 in stage 1 and 586 in stage 2); measured over active candidate jobs.
The stage-2 agent calls cost about \$1{,}700.
A candidate trains for at most 1.5 hours and, with scoring, takes at most about two hours on 4 GPUs.

\paragraph{Reuse of the development banks.}
Every candidate is scored on the same development banks, so we report gains on the hidden test bank.

\paragraph{Release.}
We will release the edit code, the development and test banks, the frozen scorer and rule, the agent prompts and stage-2 transcripts, the champion checkpoints, and the RL code with configs.
Stage-1 champions are kept; the stage-1 state file was lost.

\section{Lessons from running RSI}
\label{sec:app_lessons}

We ran the loop in two stages and a third stage that found nothing new (\cref{sec:app_rsi}).
Six lessons held across them.

\paragraph{Agent choice affects the search.}
With Claude Opus 5.5 as the agent, the loop produced eight champions and raised the hidden-test score from $0.706$ to $0.906$.
With Grok as the agent, from the same start and with the same exam and rule, it produced no new champion during the run.

\paragraph{The verifier bounds the result.}
The first stage-2 champion produces the largest step in real-video performance after the exam adds human-labeled real videos.
Through stage 1, scores on real generated videos stayed near $0.75$ ($0.700$ to $0.755$ over five champions), with the public score remaining near its starting level.
Once the exam had real videos with human labels, the first stage-2 champion raised it to $0.852$ in one step (\cref{tab:champions}).
A good exam is close to real failures, fine-grained enough to separate strong candidates, and truly hidden.
Families that the start model already solves give no signal and should be replaced by harder versions between stages.

\paragraph{Human revisions to the exam.}
Adding real videos was a human decision; public-benchmark gains followed in stage 2.
Humans also catch exam hacks that the rule allows: in stage 1, a candidate raised the smoothness mean by giving up the time-reversal family, which led to guards on single families.
The development and test banks can disagree on a family, such as time reversal or upside-down clips, so the loop should not tune to one family on the development banks alone.
Finally, humans judge the generated videos by eye.

\paragraph{Training failures account for the most unscored candidates.}
Of the 170 candidates, 52 never got a score because their training job failed, and 20 because their code was rejected (\cref{tab:outcomes}).
Smoke checks before a job starts, a hard time limit and a repair agent that reads the log reduce this waste.

\paragraph{Returns diminish.}
After a few champions, most candidates land within $\pm 0.005$ of the champion.
The 14 stage-2 candidates scored after \ours{} landed between $-0.004$ and $+0.001$, and none of the 100 candidates of the third stage met the rule.
These runs motivate expanding the exam and training data.

\paragraph{Cost of rereading the search history.}
In stage 2, half of the agent cost went to engineer sessions, because every call re-read a growing ledger (10k to 28k tokens) and the stacked champion code.
In the third stage, a rolling lessons file of at most 700 words, a short context, a code map and pruning of dead code kept the context flat for 100 candidates.

\section{Comparison protocol}
\label{sec:app_compare}

\begin{table*}[t]
\centering
\small
\setlength{\tabcolsep}{4pt}
\resizebox{\textwidth}{!}{%
\begin{tabular}{llcccccccccc}
\toprule
& & \multicolumn{2}{c}{Mean rank score} & \multicolumn{2}{c}{Dynamics} & \multicolumn{6}{c}{Selected metrics} \\
\cmidrule(lr){3-4}\cmidrule(lr){5-6}\cmidrule(lr){7-12}
Model & Type & Eval.\ (15) & Underst.\ (2) & VSB dyn. & VBench dyn. & VideoPhy-2 phys. & VideoScore-Bench & VBench pref. & T2VQA-DB & AIGVE align. & VSB-v2 text \\
\midrule
\oursrow \ours{} (ours) & open, 8B & 0.721 & 0.421 & \textbf{0.611} & \textbf{0.906} & \textbf{0.317} & \textbf{0.598} & \textbf{0.717} & \textbf{0.554} & 0.656 & 0.548 \\
After stage 1 (ours) & open, 8B & 0.434 & 0.447 & 0.564 & 0.781 & 0.092 & 0.295 & 0.707 & 0.309 & 0.518 & 0.556 \\
Qwen3-VL-8B (start) & open, 8B & 0.407 & 0.447 & 0.538 & 0.812 & 0.108 & 0.327 & 0.641 & 0.263 & 0.599 & 0.570 \\
\midrule
Qwen3-VL-32B & open & 0.542 & 0.711 & 0.495 & 0.875 & 0.156 & 0.347 & 0.685 & 0.334 & 0.630 & 0.674 \\
Qwen3.8-27B (thinking) & open & 0.628 & 0.711 & 0.453 & 0.750 & 0.221 & 0.389 & 0.601 & 0.444 & 0.696 & 0.663 \\
InternVL3.5-8B~\cite{wang2025internvl35} & open & 0.340 & 0.421 & 0.433 & 0.750 & 0.124 & 0.361 & 0.661 & 0.321 & 0.656 & 0.643 \\
VideoScore2~\cite{he2025videoscore2} & evaluator & 0.563 & 0.026 & 0.534 & 0.406 & 0.299 & 0.508 & 0.600 & 0.396 & 0.707 & \textbf{0.677} \\
UnifiedReward-7B~\cite{wang2025unifiedreward} & evaluator & 0.593 & 0.132 & 0.486 & 0.500 & 0.233 & 0.375 & 0.620 & 0.496 & 0.747 & 0.627 \\
VideoReward & evaluator & 0.425 & -- & $-$0.311 & 0.312 & 0.172 & 0.168 & 0.576 & 0.173 & 0.723 & 0.552 \\
\midrule
Claude Opus 5.5 & proprietary & \textbf{0.826} & \textbf{0.947} & 0.549 & 0.781 & 0.316 & 0.531 & 0.662 & 0.532 & \textbf{0.757} & 0.587 \\
GPT-5.6 Sol & proprietary & 0.775 & 0.895 & 0.575 & 0.844 & 0.206 & 0.521 & 0.700 & 0.530 & 0.691 & 0.626 \\
Muse Spark 1.3 & proprietary & 0.730 & 0.895 & 0.512 & 0.844 & 0.187 & 0.501 & 0.714 & 0.430 & 0.709 & 0.610 \\
Grok 4.6 & proprietary & 0.631 & 0.895 & 0.549 & 0.781 & 0.162 & 0.493 & 0.685 & 0.457 & 0.692 & 0.586 \\
Gemini 4 Argon & proprietary & 0.584 & 0.737 & 0.480 & 0.719 & 0.271 & 0.410 & 0.684 & 0.488 & 0.732 & 0.592 \\
\bottomrule
\end{tabular}}
\caption{\textbf{Comparison on 12 public benchmarks} (14 of 21 models; the other 7 are listed below).
Mean rank score: rank on each metric mapped to $[0,1]$, then averaged.
Bold: best per column.}
\label{tab:compare}
\end{table*}

\paragraph{Benchmarks.}
We compare 21 models on 12 public benchmarks: VideoPhy-2~\cite{bansal2025videophy2}, VideoScore-Bench~\cite{he2024videoscore}, VideoScore-Bench-v2~\cite{he2025videoscore2}, VBench preference pairs~\cite{huang2024vbench}, AIGVE~\cite{xiang2025aigve}, T2VQA-DB~\cite{kou2024t2vqa}, GenAI-Bench video~\cite{jiang2024genaiarena}, MJ-Bench video~\cite{tong2025mjvideo}, KoNViD-1k~\cite{hosu2017konvid}, Q-Bench-Video~\cite{zhang2024qbenchvideo}, MVBench~\cite{li2024mvbench} and VideoHallucer~\cite{wang2024videohallucer}.
They give 15 evaluation metrics and 2 understanding metrics.
Each benchmark is a fixed sample of 67 to 385 items.
The two dynamics columns of \cref{tab:compare} use the dynamics subsets: 80 VideoScore-Bench videos, and 24 VBench preference videos from 6 prompts, which give 16 pairs scored with half credit for ties.
\Cref{fig:rsi_curve}c ranks each champion alone against the other evaluators of \cref{tab:compare} except VideoReward, which changes the ranking pool relative to \cref{tab:compare}.
MVBench supplies the general guard; \cref{tab:compare} evaluates separate MVBench videos.

\paragraph{How models answer.}
Each model uses its default inference.
Open models, including \ours{} and its earlier champions, give their most likely answer; for ratings, this is the most likely integer from 1 to 5.
Proprietary models give a text answer.
VideoReward gives one reward per video, so we use it only on rating and preference benchmarks.

\paragraph{Uncertainty.}
Per-metric tests use a 90\% paired bootstrap.
Against Qwen3-VL-8B, \ours{} is significantly better on 6 metrics and worse on none.
For the mean rank score, we resample the items of each benchmark, re-rank all 21 models and average (\cref{tab:app_bench_uncertainty}).
\ours{} is above Qwen3-VL-8B in every resample and above every other open model in almost all of them; it is level with Muse Spark 1.3 and mostly below GPT-5.6 Sol and Claude Opus 5.5.

\begin{table}[ht]
\centering
\footnotesize
\setlength{\tabcolsep}{3pt}
\resizebox{\linewidth}{!}{%
\begin{tabular}{lcccc}
\toprule
Statistic & \ours{} & Claude Opus 5.5 & GPT-5.6 Sol & Muse Spark 1.3 \\
\midrule
Main result (21 models, 15 metrics) & 0.721 & 0.826 & 0.775 & 0.730 \\
\quad 90\% interval & $[0.61, 0.76]$ & $[0.71, 0.84]$ & $[0.67, 0.80]$ & $[0.61, 0.75]$ \\
\bottomrule
\end{tabular}}
\caption{\textbf{Mean rank scores and bootstrap intervals.}}
\label{tab:app_bench_uncertainty}
\end{table}

The seven models not shown in \cref{tab:compare} have these mean rank scores: Qwen3-VL-2B $0.138$, Qwen3-VL-4B $0.323$, Qwen3-VL-30B-A3B $0.419$~\cite{qwen3vl}, Qwen2.5-VL-7B $0.189$, Qwen2.5-VL-32B $0.492$~\cite{bai2025qwen25vl}, LLaVA-OneVision-7B $0.161$~\cite{li2024llavaov}, and Qwen3.8-27B without reasoning $0.572$.

\section{RL details}
\label{sec:app_rl}

\paragraph{Reward.}
The evaluator reward is a weighted sum of the dynamics, physics, smoothness and quality ratings (weights 1, 1, 1 and 0.5).
It has no text-alignment term.
An anchor loss keeps the generator close to the start model; there is no KL term.
Two penalties act on each video: one for flicker and one for global judder.
Both thresholds are above the start model's means, so the penalties act only on clear defects.
The \ours{}, Qwen3-VL-8B and stage-1 champion runs use the same reward code, penalties and settings.
All runs use the same settings, which we chose in pilot runs with \ours{}.

\paragraph{VideoReward runs.}
The fair VideoReward run sums its three scores (visual quality, motion quality and text alignment) with weight 1 each and uses the same learning rate, anchor and penalties as \ours{}.
We rescale its reward once, on the first rollout, to a within-group spread of 0.75, so that the penalties have a similar scale.
The earlier VideoReward run uses only visual and motion quality and no penalties; the figures of still videos come from this run.

\paragraph{A raw optical-flow reward.}
With one seed and the same settings, a reward of optical flow alone gives significantly more motion than \ours{} at update 50 (dynamic degree $0.932$ against $0.848$) at similar adherence.
Its videos jitter more: \ours{} is lower in local judder on both prompt sets, a measure of irregular changes of local flow that is not a penalty in either reward.
By update 100 it falls significantly behind \ours{} in both dynamic degree and adherence.
Flow rewards the amount of motion, so the generator learns jittery motion that drifts from the prompt.

\paragraph{Metrics.}
The blind judge, Qwen3.8-27B, lists the elements of each prompt and checks whether each video shows them, without knowing the source of the video.
It is from the same model family as \ours{}.
Optical flow (Farneb\"ack, on re-decoded frames) and VBench dynamic degree (RAFT~\cite{teed2020raft}) share no code or input.
Global judder, one of the two penalties, measures how irregularly the median motion of the frame changes.

\paragraph{Compute.}
\Cref{tab:rl_runs} lists the runs.
At update 50, the runs used about 70 to 80 GPU hours each.
We compare runs at matched update counts.

\begin{table}[ht]
\centering
\footnotesize
\setlength{\tabcolsep}{4pt}
\begin{tabular}{lrr}
\toprule
Reward & Updates & GPU hours \\
\midrule
\ours{}, seed 1 & 342 & 498.6 \\
\ours{}, seed 2 & 100 & 145.0 \\
Qwen3-VL-8B, seed 1 & 346 & 498.0 \\
Qwen3-VL-8B, seed 2 & 100 & 145.6 \\
Stage-1 champion & 100 & 144.4 \\
VideoReward & 130 (\cref{tab:main}: 120) & 188.7 \\
VideoReward (fair), seed 1 & 50 & 70.6 \\
VideoReward (fair), seed 2 & 50 & 70.4 \\
\bottomrule
\end{tabular}
\caption{\textbf{RL runs} from Wan2.1 T2V-1.3B. The two long runs have a budget of 500 GPU hours.}
\label{tab:rl_runs}
\end{table}

\paragraph{Choice of updates.}
Update 50 is the main comparison point, and update 100 the second; we fixed update 100 before evaluating \ours{} there.
A stopping rule based on cross-prompt similarity and a flow floor passes \ours{} and Qwen3-VL-8B at update 50, and stops VideoReward and the stage-1 champion.
We report all runs at the same updates.
The full-budget rows of \cref{tab:main} are a stress test past the stopping points.

\paragraph{Intermediate champion.}
\Cref{tab:rl_more} adds the stage-1 champion as a reward and the update-100 checkpoints.
At update 50, the stage-1 champion already gives more motion than Qwen3-VL-8B.
With seed 1, \ours{} gives about the same dynamic degree and less optical flow than the stage-1 champion; with seed 2, it gives more of both.
At update 100 on the general prompts, the gain of the stage-1 champion over Qwen3-VL-8B is no longer significant.

\begin{table}[ht]
\centering
\footnotesize
\setlength{\tabcolsep}{2.5pt}
\resizebox{\linewidth}{!}{%
\begin{tabular}{llcccc}
\toprule
Reward & Upd. & Dyn.\,$\uparrow$ & Flow\,$\uparrow$ & Adh.\,$\uparrow$ & Elem.\,$\uparrow$ \\
\midrule
Stage-1 champion & 50 & 0.857 & 18.77 & 3.659 & 0.748 \\
\midrule
Qwen3-VL-8B, seed 1 & 100 & 0.770 & 14.93 & 3.754 & 0.761 \\
Qwen3-VL-8B, seed 2 & 100 & 0.700 & 13.48 & 3.671 & 0.748 \\
Stage-1 champion & 100 & 0.803 & 15.29 & 3.648 & 0.743 \\
\ours{}, seed 1 & 100 & 0.913 & 20.53 & 3.675 & 0.750 \\
\ours{}, seed 2 & 100 & 0.892 & 17.59 & 3.519 & 0.712 \\
\bottomrule
\end{tabular}}
\caption{\textbf{More RL checkpoints} (630 test videos).}
\label{tab:rl_more}
\end{table}

\paragraph{Evaluator ratings.}
Evaluator ratings, flicker and global judder enter the training reward. We compare reward models with independent motion metrics, a separate adherence judge and human ratings.

\subsection{Full statistics of the RL comparison}
\label{sec:app_rl_stats}

\paragraph{Testing protocol.}
All differences are paired over prompts and seeds.
Intervals are 95\% bootstrap intervals from 4{,}000 resamples, and $p$-values are two-sided.
Videos of one prompt share content, so we also resample whole prompts; all confirmatory results hold under both schemes.
A difference is significant when $p<0.05$ and its interval excludes zero.
Intervals quantify variation across test prompts within a training run. We report results per seed and the gap between two Qwen3-VL-8B seeds.

\paragraph{Confirmatory tests.}
We fixed one family of 32 tests before the confirmatory analysis.
It covers VBench dynamic degree and optical flow, on both prompt sets, both seeds and updates 50 and 100, each against the start model and against Qwen3-VL-8B of the same seed and update.
All 32 tests pass after Holm correction.
In 31 of them, no resample out of 4{,}000 crosses zero.
These tests have $p<0.0005$, giving an upper bound of $0.016$ on all adjusted $p$-values.
Some bootstrap summaries of seed 1 at update 50 were already visible when we fixed this family, and we widened it from 16 to 32 tests at that point.
All other metrics are described, not tested as a family.

\paragraph{Per-seed results.}
\Cref{tab:app_rl_cells} gives every paired difference against Qwen3-VL-8B of the same seed and update.

\begin{table*}[t]
\centering
\footnotesize
\setlength{\tabcolsep}{4pt}
\begin{tabular}{ll cccc cccc}
\toprule
& & \multicolumn{4}{c}{630 test videos} & \multicolumn{4}{c}{300 general prompts} \\
\cmidrule(lr){3-6}\cmidrule(lr){7-10}
& Seed & 1 & 2 & 1 & 2 & 1 & 2 & 1 & 2 \\
& Update & 50 & 50 & 100 & 100 & 50 & 50 & 100 & 100 \\
\midrule
\multicolumn{2}{l}{\textit{Motion}\,$\uparrow$} \\
& Dynamic degree & $+0.06^*$ & $+0.14^*$ & $+0.14^*$ & $+0.19^*$ & $+0.07^*$ & $+0.10^*$ & $+0.13^*$ & $+0.14^*$ \\
& Optical flow   & $+2.0^*$ & $+4.8^*$ & $+5.6^*$ & $+4.1^*$ & $+2.7^*$ & $+6.0^*$ & $+5.4^*$ & $+4.3^*$ \\
& Object motion  & $+0.09^*$ & $+0.54^*$ & $+0.51^*$ & $+0.27^*$ & $+0.17^*$ & $+0.74^*$ & $+0.54^*$ & $+0.37^*$ \\
\midrule
\multicolumn{2}{l}{\textit{Judge}\,$\uparrow$} \\
& Adherence & $-0.01$ & $+0.03$ & $-0.08$ & $-0.15^*$ & $0.00$ & $-0.08$ & $-0.04$ & $-0.10$ \\
& Elements  & $0.00$ & $+0.01$ & $-0.01$ & $-0.04^*$ & $0.00$ & $-0.02$ & $-0.02$ & $-0.03^*$ \\
\bottomrule
\end{tabular}
\caption{\textbf{\ours{} minus Qwen3-VL-8B as the RL reward}, same seed and update. $^*$: 95\% interval excludes zero (whole prompts resampled). Dynamic degree and optical flow form the Holm family.}
\label{tab:app_rl_cells}
\end{table*}
\ours{} gives more motion in all eight cells, by all three motion measures, including object motion after camera motion is removed.
Prompt adherence does not fall significantly at update 50; it falls for seed 2 at update 100, on both prompt sets.
In the same cell, seed 2 is also below the start model in adherence on the 630 test videos, while seed 1 is level with the start model.
On the 630 videos, two Qwen3-VL-8B seeds differ in dynamic degree about as much as the \ours{} lead for seed 1 at update 50. In the other cells, the lead is at least twice this seed gap.
The automatic metrics measure the amount of motion; the human study below judges whether the videos are better.

\paragraph{Adherence over training.}
From update 50 to update 100, adherence falls in both seeds of \ours{}, more in seed 2.
Over the same updates, Qwen3-VL-8B stays roughly level.
At the full budget, \ours{} still generates motion and scores below the base model on adherence.
The Qwen3-VL-8B run degrades much more by then, so at the full budget \ours{} is ahead on adherence (one seed each).

\paragraph{Cross-prompt similarity.}
Videos for different prompts become more alike under several rewards.
At update 100, both \ours{} seeds are more alike across prompts than the Qwen3-VL-8B seed.

\paragraph{Human study.}
\label{sec:app_human}
We ran a blind study on the RL videos at update 50 (training seed 1).
Each pair shows two videos for the same prompt and sampling seed side by side, in random order and without labels.
For each pair, a rater answers four questions: which video has more motion, better visual quality and better prompt adherence, and which is better overall.
Each question also allows ``same''.
There are 48 pairs against Qwen3-VL-8B and 24 against fair VideoReward, drawn evenly from the four prompt subsets of the 630 test videos (three VBench prompt lists and VidProM).
Six raters answered all pairs.
We exclude one more answer file, which came from an earlier version of the study with data problems; including it changes no score by more than $0.02$.
The score is the share of answers for \ours{}, with ``same'' counted as half, and the 95\% intervals resample raters and pairs.

Raters prefer \ours{} on motion and overall against both references.
The motion answers follow the videos: on the 27 pairs whose optical flow differs by at least $2\times$, raters pick the video with more flow in $89\%$ of their answers that are not ``same''.
Raters also prefer \ours{} on visual quality and prompt adherence (\cref{tab:app_human}).
Against Qwen3-VL-8B, raters see more motion in the \ours{} video in 171 of 288 answers.
In these answers, they prefer \ours{} overall, in visual quality and in adherence about nine times in ten, and judge it worse overall in $3\%$.
Against fair VideoReward, they see more motion in the \ours{} video in 112 of 144 answers and prefer it overall in most of them.

\section{The RSI evaluator as a data filter}
\label{sec:app_sft}

We also tried \ours{} as a filter for fine-tuning data.
Both arms fine-tune Wan2.1 T2V-1.3B on 50k clips with the same recipe.
Data from our selection pipeline gives more motion at both steps (\cref{tab:sft} in the main paper).
The random arm first gains motion and then falls back to the start model, while the picked arm keeps its motion.
Adherence does not differ.

\paragraph{Choice of checkpoints.}
We chose the two evaluated steps after training.
We ran three picked-versus-random pairs: this pair, a smaller pair with about 20k clips per arm, whose gain in dynamic degree did not hold at the later step, and an earlier 50k pair with another recipe, which also favors selected data.

\section{More visual results}
\label{sec:app_vis}

\paragraph{Single test pairs.}
\Cref{fig:qual} shows three test pairs.
Qwen3-VL-8B rates the flawed video about the same as the good one, or even higher.
\ours{} assigns the flawed video a lower score in all three.
We selected the examples after fixing \ours{}. \Cref{tab:rsi} reports normalized scores on the full test bank.

\begin{figure*}[tp]
\centering
\includegraphics[width=\textwidth]{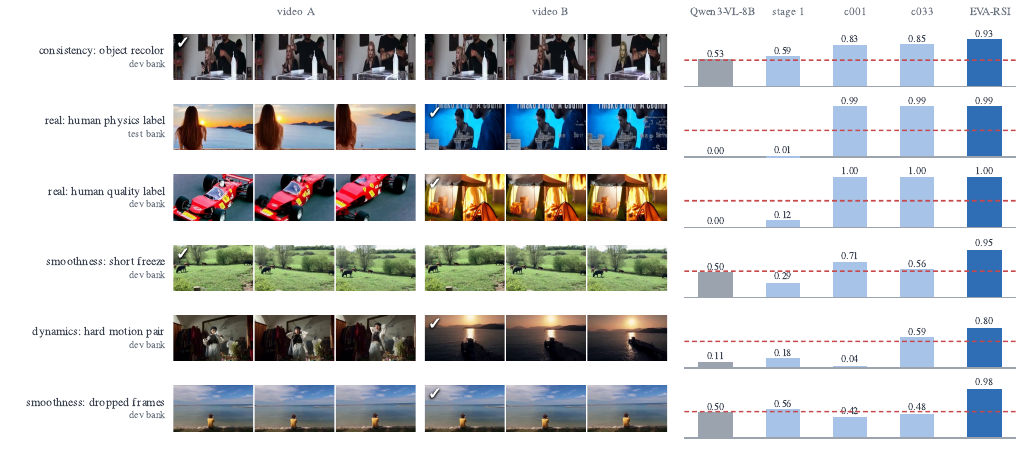}
\vspace{-5mm}
\caption{\textbf{Single questions through the chain of champions.} Bars: probability of the correct answer. \checkmark: correct video.}
\label{fig:app_rsi_evo}
\end{figure*}

\paragraph{How the evaluator changes over RSI.}
\Cref{fig:app_rsi_evo} follows single questions through the chain of champions, from Qwen3-VL-8B to \ours{}.
In every row, \ours{} gives the correct answer the highest probability, or ties with an earlier champion.
On the real-video questions, the jump comes with stage 2, which added generated videos with human labels.
On the edited questions, the last champion gives the highest probability, with its largest gains on smoothness.
None of these questions appears in the main paper.
We selected these questions after fixing \ours{}.

\begin{figure*}[tp]
\centering
\includegraphics[width=\textwidth]{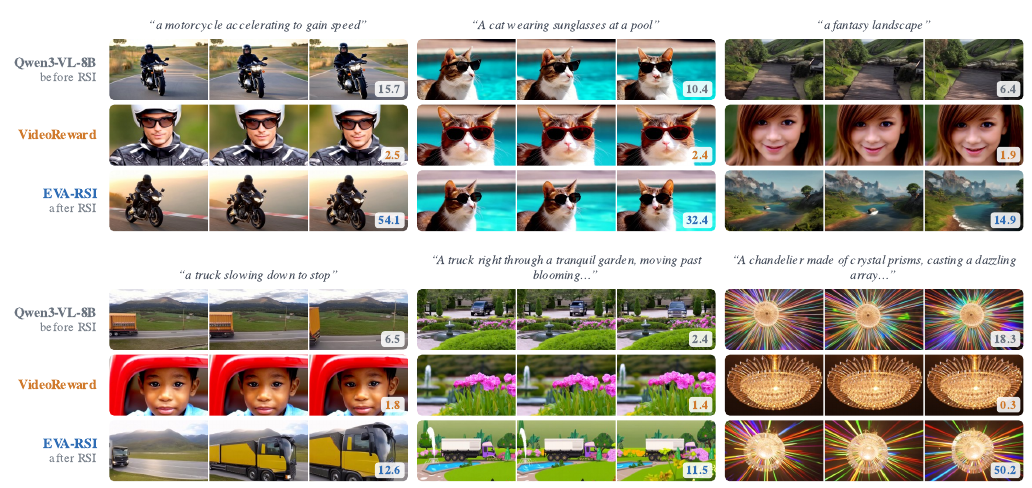}
\vspace{-5mm}
\caption{\textbf{More prompts where \ours{} moves most} (update 50). VideoReward: stress-test run.}
\label{fig:app_ds_more}
\end{figure*}

\begin{figure*}[tp]
\centering
\includegraphics[width=\textwidth]{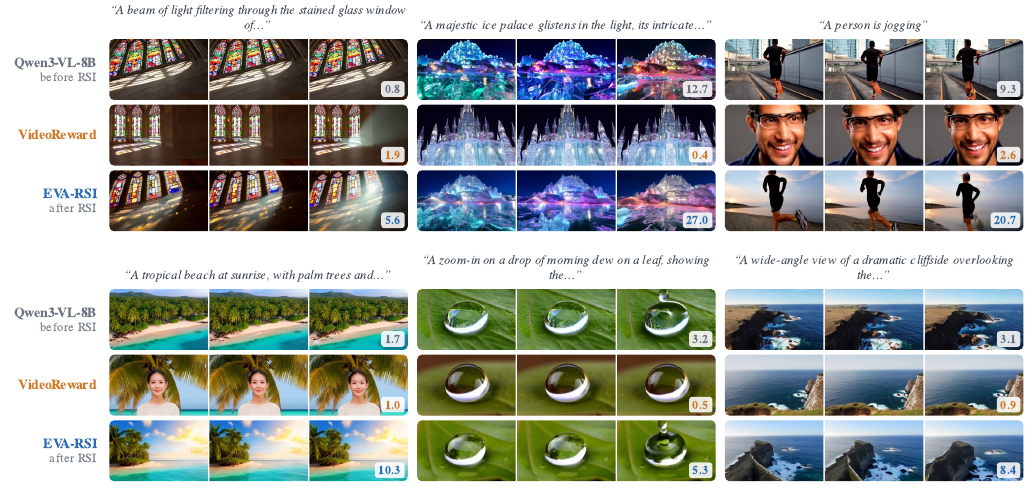}
\vspace{-5mm}
\caption{\textbf{More examples} (update 50).}
\label{fig:app_ds_pick}
\end{figure*}

\begin{figure*}[tp]
\centering
\includegraphics[width=\textwidth]{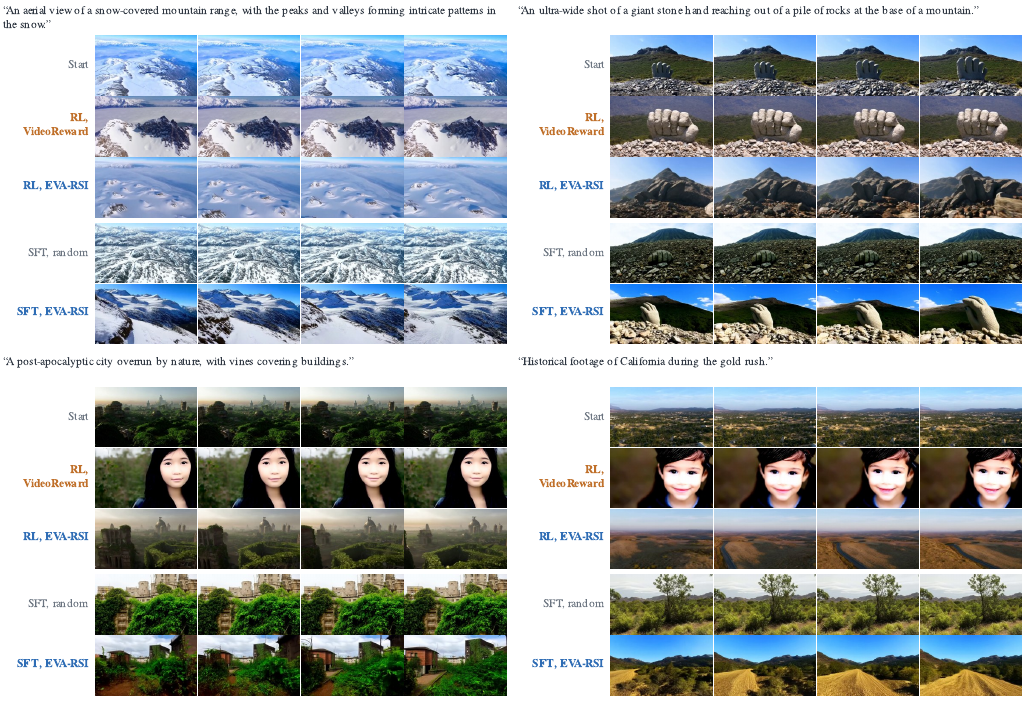}
\vspace{-5mm}
\caption{\textbf{Largest motion gains of SFT on data picked by \ours{}.}}
\label{fig:app_ds_sft_win}
\end{figure*}

\begin{figure*}[tp]
\centering
\includegraphics[width=\textwidth]{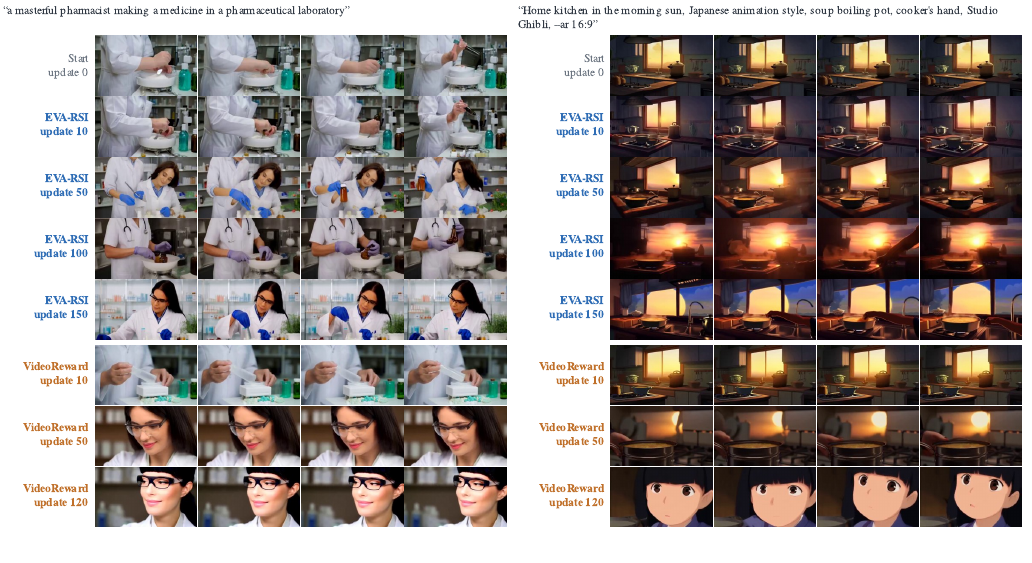}
\vspace{-5mm}
\caption{\textbf{One prompt and seed over RL training.}}
\label{fig:app_train}
\end{figure*}

\paragraph{Downstream videos.}
\Cref{fig:ds_qual} and \cref{fig:app_ds_pick,fig:app_ds_more} use the same layout: RL of Wan2.1 T2V-1.3B with three rewards, all at update 50, on the same prompt and seed.
Each cell shows three frames, and the corner number is the optical flow of the video.
We picked the examples in \cref{fig:app_ds_pick} from test prompts where \ours{} keeps adherence and moves more than Qwen3-VL-8B, and VideoReward is nearly still.
We used a minimum HPSv3~\cite{ma2025hpsv3} image-preference score solely to select these examples.
For \cref{fig:app_ds_more} we used the same rule with imaging quality in place of HPSv3: \ours{} moves at least $1.8\times$ as much as Qwen3-VL-8B, VideoReward is nearly still, and \ours{} is not worse on adherence or imaging quality.
None of these prompts appears elsewhere in the paper.
In both figures, VideoReward often turns the prompt into a still close-up of a face, and in the garden prompt it drops the truck.
Over all 630 test videos (seed 1, update 50), \ours{} has more optical flow than Qwen3-VL-8B in $57\%$, more than the fair VideoReward run in $92\%$ and more than the stress-test run in $94\%$.
\Cref{fig:app_train} follows two prompts over training, from the eight fixed videos that we log at each evaluation; its last update is past the stopping point of the run.
\Cref{fig:app_ds_sft_win} shows the four test prompts where SFT on clips picked by \ours{} is most dynamic against SFT on random clips (\cref{sec:app_sft}).

%% file: main.bib
@article{wan2025,
  author = {{Wan Team} and Wang, Ang and Ai, Baole and Wen, Bin and Mao, Chaojie and Xie, Chen-Wei and Chen, Di and Yu, Feiwu and Zhao, Haiming and Yang, Jianxiao and Zeng, Jianyuan and Wang, Jiayu and Zhang, Jingfeng and Zhou, Jingren and Wang, Jinkai and Chen, Jixuan and Zhu, Kai and Zhao, Kang and Yan, Keyu and Huang, Lianghua and Feng, Mengyang and Zhang, Ningyi and Li, Pandeng and Wu, Pingyu and Chu, Ruihang and Feng, Ruili and Zhang, Shiwei and Sun, Siyang and Fang, Tao and Wang, Tianxing and Gui, Tianyi and Weng, Tingyu and Shen, Tong and Lin, Wei and Wang, Wei and Wang, Wei and Zhou, Wenmeng and Wang, Wente and Shen, Wenting and Yu, Wenyuan and Shi, Xianzhong and Huang, Xiaoming and Xu, Xin and Kou, Yan and Lv, Yangyu and Li, Yifei and Liu, Yijing and Wang, Yiming and Zhang, Yingya and Huang, Yitong and Li, Yong and Wu, You and Liu, Yu and Pan, Yulin and Zheng, Yun and Hong, Yuntao and Shi, Yupeng and Feng, Yutong and Jiang, Zeyinzi and Han, Zhen and Wu, Zhi-Fan and Liu, Ziyu},
  title = {{Wan: Open and Advanced Large-Scale Video Generative Models}},
  journal = {arXiv preprint arXiv:2503.20314},
  year = {2025}
}

@inproceedings{liu2025videoreward,
  author = {Liu, Jie and Liu, Gongye and Liang, Jiajun and Yuan, Ziyang and Liu, Xiaokun and Zheng, Mingwu and Wu, Xiele and Wang, Qiulin and Xia, Menghan and Wang, Xintao and Liu, Xiaohong and Yang, Fei and Wan, Pengfei and ZHANG, Di and Gai, Kun and Yang, Yujiu and Ouyang, Wanli},
  title = {{Improving Video Generation with Human Feedback}},
  booktitle = {NeurIPS},
  pages = {91326--91363},
  year = {2025}
}

@inproceedings{ma2025hpsv3,
  author = {Ma, Yuhang and Wu, Xiaoshi and Sun, Keqiang and Li, Hongsheng},
  title = {{HPSv3: Towards Wide-Spectrum Human Preference Score}},
  booktitle = {ICCV},
  pages = {15086--15095},
  year = {2025}
}

@inproceedings{huang2024vbench,
  author = {Huang, Ziqi and He, Yinan and Yu, Jiashuo and Zhang, Fan and Si, Chenyang and Jiang, Yuming and Zhang, Yuanhan and Wu, Tianxing and Jin, Qingyang and Chanpaisit, Nattapol and Wang, Yaohui and Chen, Xinyuan and Wang, Limin and Lin, Dahua and Qiao, Yu and Liu, Ziwei},
  title = {{VBench: Comprehensive Benchmark Suite for Video Generative Models}},
  booktitle = {CVPR},
  pages = {21807--21818},
  year = {2024}
}

@article{qwen3vl,
  author = {Bai, Shuai and Cai, Yuxuan and Chen, Ruizhe and Chen, Keqin and Chen, Xionghui and Cheng, Zesen and Deng, Lianghao and Ding, Wei and Gao, Chang and Ge, Chunjiang and Ge, Wenbin and Guo, Zhifang and Huang, Qidong and Huang, Jie and Huang, Fei and Hui, Binyuan and Jiang, Shutong and Li, Zhaohai and Li, Mingsheng and Li, Mei and Li, Kaixin and Lin, Zicheng and Lin, Junyang and Liu, Xuejing and Liu, Jiawei and Liu, Chenglong and Liu, Yang and Liu, Dayiheng and Liu, Shixuan and Lu, Dunjie and Luo, Ruilin and Lv, Chenxu and Men, Rui and Meng, Lingchen and Ren, Xuancheng and Ren, Xingzhang and Song, Sibo and Sun, Yuchong and Tang, Jun and Tu, Jianhong and Wan, Jianqiang and Wang, Peng and Wang, Pengfei and Wang, Qiuyue and Wang, Yuxuan and Xie, Tianbao and Xu, Yiheng and Xu, Haiyang and Xu, Jin and Yang, Zhibo and Yang, Mingkun and Yang, Jianxin and Yang, An and Yu, Bowen and Zhang, Fei and Zhang, Hang and Zhang, Xi and Zheng, Bo and Zhong, Humen and Zhou, Jingren and Zhou, Fan and Zhou, Jing and Zhu, Yuanzhi and Zhu, Ke},
  title = {{Qwen3-VL Technical Report}},
  journal = {arXiv preprint arXiv:2511.21631},
  year = {2025}
}

@inproceedings{black2024ddpo,
  author = {Black, Kevin and Janner, Michael and Du, Yilun and Kostrikov, Ilya and Levine, Sergey},
  title = {{Training Diffusion Models with Reinforcement Learning}},
  booktitle = {ICLR},
  pages = {4965--4987},
  year = {2024}
}

@inproceedings{liu2025flowgrpo,
  author = {Liu, Jie and Liu, Gongye and Liang, Jiajun and Li, Yangguang and Liu, Jiaheng and Wang, Xintao and Wan, Pengfei and ZHANG, Di and Ouyang, Wanli},
  title = {{Flow-GRPO: Training Flow Matching Models via Online RL}},
  booktitle = {NeurIPS},
  pages = {45650--45685},
  year = {2025}
}

@inproceedings{wang2024vidprom,
  author = {Wang, Wenhao and Yang, Yi},
  title = {{VidProM: A Million-scale Real Prompt-Gallery Dataset for Text-to-Video Diffusion Models}},
  booktitle = {NeurIPS},
  pages = {65618--65642},
  year = {2024}
}

@inproceedings{zelikman2022star,
  author = {Zelikman, Eric and Wu, Yuhuai and Mu, Jesse and Goodman, Noah},
  title = {{STaR: Bootstrapping Reasoning With Reasoning}},
  booktitle = {NeurIPS},
  pages = {15476--15488},
  year = {2022}
}

@inproceedings{yuan2024selfrewarding,
  author = {Weizhe Yuan and Richard Yuanzhe Pang and Kyunghyun Cho and Xian Li and Sainbayar Sukhbaatar and Jing Xu and Jason E Weston},
  title = {{Self-Rewarding Language Models}},
  booktitle = {ICML},
  pages = {57905--57923},
  year = {2024}
}

@article{guo2025deepseekr1,
  author = {Guo, Daya and Yang, Dejian and Zhang, Haowei and Song, Junxiao and Wang, Peiyi and Zhu, Qihao and Xu, Runxin and Zhang, Ruoyu and Ma, Shirong and Bi, Xiao and Zhang, Xiaokang and Yu, Xingkai and Wu, Yu and Wu, Z. F. and Gou, Zhibin and Shao, Zhihong and Li, Zhuoshu and Gao, Ziyi and Liu, Aixin and Xue, Bing and Wang, Bingxuan and Wu, Bochao and Feng, Bei and Lu, Chengda and Zhao, Chenggang and Deng, Chengqi and Ruan, Chong and Dai, Damai and Chen, Deli and Ji, Dongjie and Li, Erhang and Lin, Fangyun and Dai, Fucong and Luo, Fuli and Hao, Guangbo and Chen, Guanting and Li, Guowei and Zhang, H. and Xu, Hanwei and Ding, Honghui and Gao, Huazuo and Qu, Hui and Li, Hui and Guo, Jianzhong and Li, Jiashi and Chen, Jingchang and Yuan, Jingyang and Tu, Jinhao and Qiu, Junjie and Li, Junlong and Cai, J. L. and Ni, Jiaqi and Liang, Jian and Chen, Jin and Dong, Kai and Hu, Kai and You, Kaichao and Gao, Kaige and Guan, Kang and Huang, Kexin and Yu, Kuai and Wang, Lean and Zhang, Lecong and Zhao, Liang and Wang, Litong and Zhang, Liyue and Xu, Lei and Xia, Leyi and Zhang, Mingchuan and Zhang, Minghua and Tang, Minghui and Zhou, Mingxu and Li, Meng and Wang, Miaojun and Li, Mingming and Tian, Ning and Huang, Panpan and Zhang, Peng and Wang, Qiancheng and Chen, Qinyu and Du, Qiushi and Ge, Ruiqi and Zhang, Ruisong and Pan, Ruizhe and Wang, Runji and Chen, R. J. and Jin, R. L. and Chen, Ruyi and Lu, Shanghao and Zhou, Shangyan and Chen, Shanhuang and Ye, Shengfeng and Wang, Shiyu and Yu, Shuiping and Zhou, Shunfeng and Pan, Shuting and Li, S. S. and Zhou, Shuang and Wu, Shaoqing and Yun, Tao and Pei, Tian and Sun, Tianyu and Wang, T. and Zeng, Wangding and Liu, Wen and Liang, Wenfeng and Gao, Wenjun and Yu, Wenqin and Zhang, Wentao and Xiao, W. L. and An, Wei and Liu, Xiaodong and Wang, Xiaohan and Chen, Xiaokang and Nie, Xiaotao and Cheng, Xin and Liu, Xin and Xie, Xin and Liu, Xingchao and Yang, Xinyu and Li, Xinyuan and Su, Xuecheng and Lin, Xuheng and Li, X. Q. and Jin, Xiangyue and Shen, Xiaojin and Chen, Xiaosha and Sun, Xiaowen and Wang, Xiaoxiang and Song, Xinnan and Zhou, Xinyi and Wang, Xianzu and Shan, Xinxia and Li, Y. K. and Wang, Y. Q. and Wei, Y. X. and Zhang, Yang and Xu, Yanhong and Li, Yao and Zhao, Yao and Sun, Yaofeng and Wang, Yaohui and Yu, Yi and Zhang, Yichao and Shi, Yifan and Xiong, Yiliang and He, Ying and Piao, Yishi and Wang, Yisong and Tan, Yixuan and Ma, Yiyang and Liu, Yiyuan and Guo, Yongqiang and Ou, Yuan and Wang, Yuduan and Gong, Yue and Zou, Yuheng and He, Yujia and Xiong, Yunfan and Luo, Yuxiang and You, Yuxiang and Liu, Yuxuan and Zhou, Yuyang and Zhu, Y. X. and Huang, Yanping and Li, Yaohui and Zheng, Yi and Zhu, Yuchen and Ma, Yunxian and Tang, Ying and Zha, Yukun and Yan, Yuting and Ren, Z. Z. and Ren, Zehui and Sha, Zhangli and Fu, Zhe and Xu, Zhean and Xie, Zhenda and Zhang, Zhengyan and Hao, Zhewen and Ma, Zhicheng and Yan, Zhigang and Wu, Zhiyu and Gu, Zihui and Zhu, Zijia and Liu, Zijun and Li, Zilin and Xie, Ziwei and Song, Ziyang and Pan, Zizheng and Huang, Zhen and Xu, Zhipeng and Zhang, Zhongyu and Zhang, Zhen},
  title = {{DeepSeek-R1 incentivizes reasoning in LLMs through reinforcement learning}},
  journal = {Nature},
  volume = {645},
  number = {8081},
  pages = {633--638},
  year = {2025}
}

@inproceedings{lambert2024tulu3,
  author = {Lambert, Nathan and Morrison, Jacob and Pyatkin, Valentina and Huang, Shengyi and Ivison, Hamish and Brahman, Faeze and Miranda, Lester James and Liu, Alisa and Dziri, Nouha and Lyu, Xinxi and Gu, Yuling and Malik, Saumya and Graf, Victoria and Hwang, Jena and Yang, Jiangjiang and Le Bras, Ronan and Tafjord, Oyvind and Wilhelm, Christopher and Soldaini, Luca and Smith, Noah and Wang, Yizhong and Dasigi, Pradeep and Hajishirzi, Hanna},
  title = {{Tulu 3: Pushing Frontiers in Open Language Model Post-Training}},
  booktitle = {COLM},
  year = {2025}
}

@article{nash1950bargaining,
  author = {Nash, Jr., John F.},
  title = {{The Bargaining Problem}},
  journal = {Econometrica},
  volume = {18},
  number = {2},
  pages = {155--162},
  year = {1950}
}

@inproceedings{bansal2025videophy2,
  author = {Bansal, Hritik and Peng, Clark and Bitton, Yonatan and Goldenberg, Roman and Grover, Aditya and Chang, Kai-Wei},
  title = {{VideoPhy-2: A Challenging Action-Centric Physical Commonsense Evaluation in Video Generation}},
  booktitle = {ICLR},
  pages = {118456--118470},
  year = {2026}
}

@inproceedings{he2024videoscore,
  author = {He, Xuan and Jiang, Dongfu and Zhang, Ge and Ku, Max and Soni, Achint and Siu, Sherman and Chen, Haonan and Chandra, Abhranil and Jiang, Ziyan and Arulraj, Aaran and Wang, Kai and Do, Quy Duc and Ni, Yuansheng and Lyu, Bohan and Narsupalli, Yaswanth and Fan, Rongqi and Lyu, Zhiheng and Lin, Bill Yuchen and Chen, Wenhu},
  title = {{VideoScore: Building Automatic Metrics to Simulate Fine-grained Human Feedback for Video Generation}},
  booktitle = {EMNLP},
  pages = {2105--2123},
  year = {2024}
}

@inproceedings{li2024mvbench,
  author = {Li, Kunchang and Wang, Yali and He, Yinan and Li, Yizhuo and Wang, Yi and Liu, Yi and Wang, Zun and Xu, Jilan and Chen, Guo and Luo, Ping and Wang, Limin and Qiao, Yu},
  title = {{MVBench: A Comprehensive Multi-modal Video Understanding Benchmark}},
  booktitle = {CVPR},
  pages = {22195--22206},
  year = {2024}
}

@inproceedings{liu2024tempcompass,
  author = {Liu, Yuanxin and Li, Shicheng and Liu, Yi and Wang, Yuxiang and Ren, Shuhuai and Li, Lei and Chen, Sishuo and Sun, Xu and Hou, Lu},
  title = {{TempCompass: Do Video LLMs Really Understand Videos?}},
  booktitle = {Findings of ACL},
  pages = {8731--8772},
  year = {2024}
}

@article{diffusionart,
  author = {Wan, Zhenglin and Chen, Jiayu and Zhang, Zhengfeiyang and Zhou, Pengfei and Zhao, Wangbo and Kong, Fangyuan and Wan, Pengfei and Gai, Kun and You, Yang},
  title = {{DiffusionART: Efficient, Stable and Scalable Reinforcement of Diffusion Models}},
  journal = {arXiv preprint arXiv:2610.12572},
  year = {2026}
}

@article{blattmann2023svd,
  author = {Blattmann, Andreas and Dockhorn, Tim and Kulal, Sumith and Mendelevitch, Daniel and Kilian, Maciej and Lorenz, Dominik and Levi, Yam and English, Zion and Voleti, Vikram and Letts, Adam and Jampani, Varun and Rombach, Robin},
  title = {{Stable Video Diffusion: Scaling Latent Video Diffusion Models to Large Datasets}},
  journal = {arXiv preprint arXiv:2311.15127},
  year = {2023}
}

@inproceedings{nan2025openvid,
  author = {Nan, Kepan and Xie, Rui and Zhou, Penghao and Fan, Tiehan and Yang, Zhenheng and Chen, Zhijie and Li, Xiang and Yang, Jian and Tai, Ying},
  title = {{OpenVid-1M: A Large-Scale High-Quality Dataset for Text-to-video Generation}},
  booktitle = {ICLR},
  pages = {1045--1064},
  year = {2025}
}

@article{zhang2026flexrsi,
  author = {Zhang, Jingdong and Wang, Jionghao and Yu, Zhengming and Cai, Yancheng and Chow, Wei and Choy, Chris and Dou, Zhiyang and Hua, Hang and Li, Jiachen and Liu, Zhixuan and Wang, Chen and Wei, Songlin and Yang, Siyuan and Yuan, Yu and Zhan, Xiaohang and Zhang, Yucheng and Zheng, Shuhong and Li, Xin and Liu, Lingjie and Matusik, Wojciech and Tu, Zhengzhong and Wang, Wenping},
  title = {{Flex-RSI: Benchmarking Recursive Self-Improvement of Frontier Agents}},
  journal = {arXiv preprint arXiv:2611.08946},
  year = {2026}
}

@article{zheng2026dreamrsi,
  author = {Zheng, Tong and Wu, Xidong and Zhang, Zheng and He, Zhankui and Zhang, Chaoyi and Coleman, Benjamin and Wei, Ruoqiao and Bai, Di and Liu, Haolin and Liu, Rui and Wang, Xue and Zhuan, Yue and Kang, Wang-Cheng and Xiang, Renkai and Huang, Heng and Cheng, Xinwu and Guo, Yunsong},
  title = {{Dream-RSI: Recursive Self-Improvement through Evolving Worlds}},
  journal = {arXiv preprint arXiv:2609.14858},
  year = {2026}
}

@inproceedings{zhang2026dgm,
  author = {Zhang, Jenny and Hu, Shengran and Lu, Cong and Lange, Robert and Clune, Jeff},
  title = {{Darwin G{\"o}del Machine: Open-Ended Evolution of Self-Improving Agents}},
  booktitle = {ICLR},
  year = {2026}
}

@inproceedings{lange2026shinka,
  author = {Lange, Robert and Imajuku, Yuki and Cetin, Edoardo},
  title = {{ShinkaEvolve: Towards Open-Ended and Sample-Efficient Program Evolution}},
  booktitle = {ICLR},
  pages = {74026--74078},
  year = {2026}
}

@article{novikov2025alphaevolve,
  author = {Novikov, Alexander and V{\~u}, Ng{\^a}n and Eisenberger, Marvin and Dupont, Emilien and Huang, Po-Sen and Wagner, Adam Zsolt and Shirobokov, Sergey and Kozlovskii, Borislav and Ruiz, Francisco J. R. and Mehrabian, Abbas and Kumar, M. Pawan and See, Abigail and Chaudhuri, Swarat and Holland, George and Davies, Alex and Nowozin, Sebastian and Kohli, Pushmeet and Balog, Matej},
  title = {{AlphaEvolve: A coding agent for scientific and algorithmic discovery}},
  journal = {arXiv preprint arXiv:2506.13131},
  year = {2025}
}

@article{romera2024funsearch,
  author = {Romera-Paredes, Bernardino and Barekatain, Mohammadamin and Novikov, Alexander and Balog, Matej and Kumar, M. Pawan and Dupont, Emilien and Ruiz, Francisco J. R. and Ellenberg, Jordan S. and Wang, Pengming and Fawzi, Omar and Kohli, Pushmeet and Fawzi, Alhussein},
  title = {{Mathematical discoveries from program search with large language models}},
  journal = {Nature},
  volume = {625},
  number = {7995},
  pages = {468--475},
  year = {2024}
}

@inproceedings{hu2025adas,
  author = {Hu, Shengran and Lu, Cong and Clune, Jeff},
  title = {{Automated Design of Agentic Systems}},
  booktitle = {ICLR},
  pages = {21344--21377},
  year = {2025}
}

@inproceedings{zelikman2024stop,
  author = {Zelikman, Eric and Lorch, Eliana and Mackey, Lester and Kalai, Adam T.},
  title = {{Self-Taught Optimizer (STOP): Recursively Self-Improving Code Generation}},
  booktitle = {COLM},
  year = {2024}
}

@article{lu2024aiscientist,
  author = {Lu, Chris and Lu, Cong and Lange, Robert Tjarko and Foerster, Jakob and Clune, Jeff and Ha, David},
  title = {{The AI Scientist: Towards Fully Automated Open-Ended Scientific Discovery}},
  journal = {arXiv preprint arXiv:2408.06292},
  year = {2024}
}

@inproceedings{chan2025mlebench,
  author = {Chan, Jun Shern and Chowdhury, Neil and Jaffe, Oliver and Aung, James and Sherburn, Dane and Mays, Evan and Starace, Giulio and Liu, Kevin and Maksin, Leon and Patwardhan, Tejal and Madry, Aleksander and Weng, Lilian},
  title = {{MLE-bench: Evaluating Machine Learning Agents on Machine Learning Engineering}},
  booktitle = {ICLR},
  pages = {50466--50494},
  year = {2025}
}

@inproceedings{yang2024opro,
  author = {Yang, Chengrun and Wang, Xuezhi and Lu, Yifeng and Liu, Hanxiao and Le, Quoc V and Zhou, Denny and Chen, Xinyun},
  title = {{Large Language Models as Optimizers}},
  booktitle = {ICLR},
  pages = {12028--12068},
  year = {2024}
}

@inproceedings{huang2023selfimprove,
  author = {Jiaxin Huang and Shixiang Gu and Le Hou and Yuexin Wu and Xuezhi Wang and Hongkun Yu and Jiawei Han},
  title = {{Large Language Models Can Self-Improve}},
  booktitle = {EMNLP},
  pages = {1051--1068},
  year = {2023}
}

@inproceedings{madaan2023selfrefine,
  author = {Madaan, Aman and Tandon, Niket and Gupta, Prakhar and Hallinan, Skyler and Gao, Luyu and Wiegreffe, Sarah and Alon, Uri and Dziri, Nouha and Prabhumoye, Shrimai and Yang, Yiming and Gupta, Shashank and Majumder, Bodhisattwa Prasad and Hermann, Katherine and Welleck, Sean and Yazdanbakhsh, Amir and Clark, Peter},
  title = {{Self-Refine: Iterative Refinement with Self-Feedback}},
  booktitle = {NeurIPS},
  pages = {46534--46594},
  year = {2023}
}

@article{gulcehre2023rest,
  author = {Gulcehre, Caglar and Paine, Tom Le and Srinivasan, Srivatsan and Konyushkova, Ksenia and Weerts, Lotte and Sharma, Abhishek and Siddhant, Aditya and Ahern, Alex and Wang, Miaosen and Gu, Chenjie and Macherey, Wolfgang and Doucet, Arnaud and Firat, Orhan and de Freitas, Nando},
  title = {{Reinforced Self-Training (ReST) for Language Modeling}},
  journal = {arXiv preprint arXiv:2308.08998},
  year = {2023}
}

@inproceedings{liu2024evalcrafter,
  author = {Liu, Yaofang and Cun, Xiaodong and Liu, Xuebo and Wang, Xintao and Zhang, Yong and Chen, Haoxin and Liu, Yang and Zeng, Tieyong and Chan, Raymond and Shan, Ying},
  title = {{EvalCrafter: Benchmarking and Evaluating Large Video Generation Models}},
  booktitle = {CVPR},
  pages = {22139--22149},
  year = {2024}
}

@inproceedings{sun2025t2vcompbench,
  author = {Sun, Kaiyue and Huang, Kaiyi and Liu, Xian and Wu, Yue and Xu, Zihan and Li, Zhenguo and Liu, Xihui},
  title = {{T2V-CompBench: A Comprehensive Benchmark for Compositional Text-to-video Generation}},
  booktitle = {CVPR},
  pages = {8406--8416},
  year = {2025}
}

@inproceedings{bansal2025videophy,
  author = {Bansal, Hritik and Lin, Zongyu and Xie, Tianyi and Zong, Zeshun and Yarom, Michal and Bitton, Yonatan and Jiang, Chenfanfu and Sun, Yizhou and Chang, Kai-Wei and Grover, Aditya},
  title = {{VideoPhy: Evaluating Physical Commonsense for Video Generation}},
  booktitle = {ICLR},
  pages = {102075--102121},
  year = {2025}
}

@inproceedings{xu2024visionreward,
  author = {Xu, Jiazheng and Huang, Yu and Cheng, Jiale and Yang, Yuanming and Xu, Jiajun and Wang, Yuan and Duan, Wenbo and Yang, Shen and Jin, Qunlin and Li, Shurun and Teng, Jiayan and Yang, Zhuoyi and Zheng, Wendi and Liu, Xiao and Zhang, Dan and Ding, Ming and Zhang, Xiaohan and Huang, Shiyu and Gu, Xiaotao and Huang, Minlie and Tang, Jie and Dong, Yuxiao},
  title = {{VisionReward: Fine-Grained Multi-Dimensional Human Preference Learning for Image and Video Generation}},
  booktitle = {AAAI},
  pages = {11269--11277},
  year = {2026}
}

@article{wang2025unifiedreward,
  author = {Wang, Yibin and Zang, Yuhang and Li, Hao and Jin, Cheng and Wang, Jiaqi},
  title = {{Unified Reward Model for Multimodal Understanding and Generation}},
  journal = {arXiv preprint arXiv:2503.05236},
  year = {2025}
}

@inproceedings{zheng2023judging,
  author = {Zheng, Lianmin and Chiang, Wei-Lin and Sheng, Ying and Zhuang, Siyuan and Wu, Zhanghao and Zhuang, Yonghao and Lin, Zi and Li, Zhuohan and Li, Dacheng and Xing, Eric and Zhang, Hao and Gonzalez, Joseph and Stoica, Ion},
  title = {{Judging LLM-as-a-Judge with MT-Bench and Chatbot Arena}},
  booktitle = {NeurIPS},
  pages = {46595--46623},
  year = {2023}
}

@inproceedings{fu2025videomme,
  author = {Fu, Chaoyou and Dai, Yuhan and Luo, Yongdong and Li, Lei and Ren, Shuhuai and Zhang, Renrui and Wang, Zihan and Zhou, Chenyu and Shen, Yunhang and Zhang, Mengdan and Chen, Peixian and Li, Yanwei and Lin, Shaohui and Zhao, Sirui and Li, Ke and Xu, Tong and Zheng, Xiawu and Chen, Enhong and Shan, Caifeng and He, Ran and Sun, Xing},
  title = {{Video-MME: The First-Ever Comprehensive Evaluation Benchmark of Multi-modal LLMs in Video Analysis}},
  booktitle = {CVPR},
  pages = {24108--24118},
  year = {2025}
}

@inproceedings{liu2017rankiqa,
  author = {Liu, Xialei and Van De Weijer, Joost and Bagdanov, Andrew D.},
  title = {{RankIQA: Learning from Rankings for No-Reference Image Quality Assessment}},
  booktitle = {ICCV},
  pages = {1040--1049},
  year = {2017}
}

@inproceedings{lin2019kadid,
  author = {Lin, Hanhe and Hosu, Vlad and Saupe, Dietmar},
  title = {{KADID-10k: A Large-scale Artificially Distorted IQA Database}},
  booktitle = {QoMEX},
  pages = {1--3},
  year = {2019}
}

@inproceedings{bagad2023testoftime,
  author = {Bagad, Piyush and Tapaswi, Makarand and Snoek, Cees G.M.},
  title = {{Test of Time: Instilling Video-Language Models with a Sense of Time}},
  booktitle = {CVPR},
  pages = {2503--2516},
  year = {2023}
}

@article{he2025videoscore2,
  author = {Xuan He and Dongfu Jiang and Ping Nie and Minghao Liu and Zhengxuan Jiang and Mingyi Su and Wentao Ma and Junru Lin and Chun Ye and Yi Lu and Keming Wu and Benjamin Schneider and Quy Duc Do and Zhuofeng Li and Yiming Jia and Yuxuan Zhang and Guo Cheng and Haozhe Wang and Wangchunshu Zhou and Qunshu Lin and Yuanxing Zhang and Ge Zhang and Wenhao Huang and Wenhu Chen},
  title = {{VideoScore2: Think Before You Score In Generated Video Evaluation}},
  journal = {Transactions on Machine Learning Research},
  year = {2026}
}

@article{wang2025internvl35,
  author = {Wang, Weiyun and Gao, Zhangwei and Gu, Lixin and Pu, Hengjun and Cui, Long and Wei, Xingguang and Liu, Zhaoyang and Jing, Linglin and Ye, Shenglong and Shao, Jie and Wang, Zhaokai and Chen, Zhe and Zhang, Hongjie and Yang, Ganlin and Wang, Haomin and Wei, Qi and Yin, Jinhui and Li, Wenhao and Cui, Erfei and Chen, Guanzhou and Ding, Zichen and Tian, Changyao and Wu, Zhenyu and Xie, Jingjing and Li, Zehao and Yang, Bowen and Duan, Yuchen and Wang, Xuehui and Hou, Zhi and Hao, Haoran and Zhang, Tianyi and Li, Songze and Zhao, Xiangyu and Duan, Haodong and Deng, Nianchen and Fu, Bin and He, Yinan and Wang, Yi and He, Conghui and Shi, Botian and He, Junjun and Xiong, Yingtong and Lv, Han and Wu, Lijun and Shao, Wenqi and Zhang, Kaipeng and Deng, Huipeng and Qi, Biqing and Ge, Jiaye and Guo, Qipeng and Zhang, Wenwei and Zhang, Songyang and Cao, Maosong and Lin, Junyao and Tang, Kexian and Gao, Jianfei and Huang, Haian and Gu, Yuzhe and Lyu, Chengqi and Tang, Huanze and Wang, Rui and Lv, Haijun and Ouyang, Wanli and Wang, Limin and Dou, Min and Zhu, Xizhou and Lu, Tong and Lin, Dahua and Dai, Jifeng and Su, Weijie and Zhou, Bowen and Chen, Kai and Qiao, Yu and Wang, Wenhai and Luo, Gen},
  title = {{InternVL3.5: Advancing Open-Source Multimodal Models in Versatility, Reasoning, and Efficiency}},
  journal = {arXiv preprint arXiv:2508.18265},
  year = {2025}
}

@article{bai2025qwen25vl,
  author = {Bai, Shuai and Chen, Keqin and Liu, Xuejing and Wang, Jialin and Ge, Wenbin and Song, Sibo and Dang, Kai and Wang, Peng and Wang, Shijie and Tang, Jun and Zhong, Humen and Zhu, Yuanzhi and Yang, Mingkun and Li, Zhaohai and Wan, Jianqiang and Wang, Pengfei and Ding, Wei and Fu, Zheren and Xu, Yiheng and Ye, Jiabo and Zhang, Xi and Xie, Tianbao and Cheng, Zesen and Zhang, Hang and Yang, Zhibo and Xu, Haiyang and Lin, Junyang},
  title = {{Qwen2.5-VL Technical Report}},
  journal = {arXiv preprint arXiv:2502.13923},
  year = {2025}
}

@article{li2024llavaov,
  author = {Bo Li and Yuanhan Zhang and Dong Guo and Renrui Zhang and Feng Li and Hao Zhang and Kaichen Zhang and Peiyuan Zhang and Yanwei Li and Ziwei Liu and Chunyuan Li},
  title = {{LLaVA-OneVision: Easy Visual Task Transfer}},
  journal = {Transactions on Machine Learning Research},
  year = {2025}
}

@inproceedings{teed2020raft,
  author = {Teed, Zachary and Deng, Jia},
  title = {{RAFT: Recurrent All-Pairs Field Transforms for Optical Flow}},
  booktitle = {ECCV},
  pages = {402--419},
  year = {2020}
}

@article{polyak2024moviegen,
  author = {Polyak, Adam and Zohar, Amit and Brown, Andrew and Tjandra, Andros and Sinha, Animesh and Lee, Ann and Vyas, Apoorv and Shi, Bowen and Ma, Chih-Yao and Chuang, Ching-Yao and Yan, David and Choudhary, Dhruv and Wang, Dingkang and Sethi, Geet and Pang, Guan and Ma, Haoyu and Misra, Ishan and Hou, Ji and Wang, Jialiang and Jagadeesh, Kiran and Li, Kunpeng and Zhang, Luxin and Singh, Mannat and Williamson, Mary and Le, Matt and Yu, Matthew and Singh, Mitesh Kumar and Zhang, Peizhao and Vajda, Peter and Duval, Quentin and Girdhar, Rohit and Sumbaly, Roshan and Rambhatla, Sai Saketh and Tsai, Sam and Azadi, Samaneh and Datta, Samyak and Chen, Sanyuan and Bell, Sean and Ramaswamy, Sharadh and Sheynin, Shelly and Bhattacharya, Siddharth and Motwani, Simran and Xu, Tao and Li, Tianhe and Hou, Tingbo and Hsu, Wei-Ning and Yin, Xi and Dai, Xiaoliang and Taigman, Yaniv and Luo, Yaqiao and Liu, Yen-Cheng and Wu, Yi-Chiao and Zhao, Yue and Kirstain, Yuval and He, Zecheng and He, Zijian and Pumarola, Albert and Thabet, Ali and Sanakoyeu, Artsiom and Mallya, Arun and Guo, Baishan and Araya, Boris and Kerr, Breena and Wood, Carleigh and Liu, Ce and Peng, Cen and Vengertsev, Dimitry and Schonfeld, Edgar and Blanchard, Elliot and Juefei-Xu, Felix and Nord, Fraylie and Liang, Jeff and Hoffman, John and Kohler, Jonas and Fire, Kaolin and Sivakumar, Karthik and Chen, Lawrence and Yu, Licheng and Gao, Luya and Georgopoulos, Markos and Moritz, Rashel and Sampson, Sara K. and Li, Shikai and Parmeggiani, Simone and Fine, Steve and Fowler, Tara and Petrovic, Vladan and Du, Yuming},
  title = {{Movie Gen: A Cast of Media Foundation Models}},
  journal = {arXiv preprint arXiv:2410.13720},
  year = {2024}
}

@inproceedings{wu2022fastvqa,
  author = {Wu, Haoning and Chen, Chaofeng and Hou, Jingwen and Liao, Liang and Wang, Annan and Sun, Wenxiu and Yan, Qiong and Lin, Weisi},
  title = {{FAST-VQA: Efficient End-to-End Video Quality Assessment with Fragment Sampling}},
  booktitle = {ECCV},
  pages = {538--554},
  year = {2022}
}

@inproceedings{wu2023dover,
  author = {Wu, Haoning and Zhang, Erli and Liao, Liang and Chen, Chaofeng and Hou, Jingwen and Wang, Annan and Sun, Wenxiu and Yan, Qiong and Lin, Weisi},
  title = {{Exploring Video Quality Assessment on User Generated Contents from Aesthetic and Technical Perspectives}},
  booktitle = {ICCV},
  pages = {20087--20097},
  year = {2023}
}

@inproceedings{wu2024qalign,
  author = {Haoning Wu and Zicheng Zhang and Weixia Zhang and Chaofeng Chen and Liang Liao and Chunyi Li and Yixuan Gao and Annan Wang and Erli Zhang and Wenxiu Sun and Qiong Yan and Xiongkuo Min and Guangtao Zhai and Weisi Lin},
  title = {{Q-Align: Teaching LMMs for Visual Scoring via Discrete Text-Defined Levels}},
  booktitle = {ICML},
  pages = {54015--54029},
  year = {2024}
}

@inproceedings{wu2024qinstruct,
  author = {Wu, Haoning and Zhang, Zicheng and Zhang, Erli and Chen, Chaofeng and Liao, Liang and Wang, Annan and Xu, Kaixin and Li, Chunyi and Hou, Jingwen and Zhai, Guangtao and Xue, Geng and Sun, Wenxiu and Yan, Qiong and Lin, Weisi},
  title = {{Q-Instruct: Improving Low-level Visual Abilities for Multi-modality Foundation Models}},
  booktitle = {CVPR},
  pages = {25490--25500},
  year = {2024}
}

@inproceedings{you2024depictqa,
  author = {You, Zhiyuan and Li, Zheyuan and Gu, Jinjin and Yin, Zhenfei and Xue, Tianfan and Dong, Chao},
  title = {{Depicting Beyond Scores: Advancing Image Quality Assessment Through Multi-modal Language Models}},
  booktitle = {ECCV},
  pages = {259--276},
  year = {2024}
}

@inproceedings{kou2024t2vqa,
  author = {Kou, Tengchuan and Liu, Xiaohong and Zhang, Zicheng and Li, Chunyi and Wu, Haoning and Min, Xiongkuo and Zhai, Guangtao and Liu, Ning},
  title = {{Subjective-Aligned Dataset and Metric for Text-to-Video Quality Assessment}},
  booktitle = {ACM Multimedia},
  pages = {7793--7802},
  year = {2024}
}

@article{wang2024lift,
  author = {Wang, Yibin and Tan, Zhiyu and Wang, Junyan and Yang, Xiaomeng and Jin, Cheng and Li, Hao},
  title = {{LiFT: Leveraging Human Feedback for Text-to-Video Model Alignment}},
  journal = {arXiv preprint arXiv:2412.04814},
  year = {2024}
}

@inproceedings{hosu2017konvid,
  author = {Hosu, Vlad and Hahn, Franz and Jenadeleh, Mohsen and Lin, Hanhe and Men, Hui and Sziranyi, Tamas and Li, Shujun and Saupe, Dietmar},
  title = {{The Konstanz natural video database (KoNViD-1k)}},
  booktitle = {QoMEX},
  pages = {1--6},
  year = {2017}
}

@inproceedings{zhang2024qbenchvideo,
  author = {Zhang, Zicheng and Jia, Ziheng and Wu, Haoning and Li, Chunyi and Chen, Zijian and Zhou, Yingjie and Sun, Wei and Liu, Xiaohong and Min, Xiongkuo and Lin, Weisi and Zhai, Guangtao},
  title = {{Q-Bench-Video: Benchmark the Video Quality Understanding of LMMs}},
  booktitle = {CVPR},
  pages = {3229--3239},
  year = {2025}
}

@article{wang2024videohallucer,
  author = {Wang, Yuxuan and Wang, Yueqian and Zhao, Dongyan and Xie, Cihang and Zheng, Zilong},
  title = {{VideoHallucer: Evaluating Intrinsic and Extrinsic Hallucinations in Large Video-Language Models}},
  journal = {arXiv preprint arXiv:2406.16338},
  year = {2024}
}

@inproceedings{jiang2024genaiarena,
  author = {Jiang, Dongfu and Ku, Max and Li, Tianle and Ni, Yuansheng and Sun, Shizhuo and Fan, Rongqi and Chen, Wenhu},
  title = {{GenAI Arena: An Open Evaluation Platform for Generative Models}},
  booktitle = {NeurIPS},
  pages = {79889--79908},
  year = {2024}
}

@inproceedings{tong2025mjvideo,
  author = {Tong, Haibo and Wang, Zhaoyang and Chen, Zhaorun and Ji, Haonian and Qiu, Shi and Han, Siwei and Geng, Kexin and Xue, Zhongkai and Zhou, Yiyang and Xia, Peng and Ding, Mingyu and Rafailov, Rafael and Finn, Chelsea and Yao, Huaxiu},
  title = {{MJ-Video: Benchmarking and Rewarding Video Generation with Fine-Grained Video Preference}},
  booktitle = {NeurIPS},
  pages = {79142--79173},
  year = {2025}
}

@inproceedings{xiang2025aigve,
  author = {Xiang, Xinhao and Liu, Xiao and Li, Zizhong and Liu, Zhuosheng and Zhang, Jiawei},
  title = {{AIGVE-Tool: AI-Generated Video Evaluation Toolkit with Multifaceted Benchmark}},
  booktitle = {VISAPP},
  pages = {346--357},
  year = {2026}
}

@inproceedings{wang2025motioncritic,
  author = {Wang, Haoru and Zhu, Wentao and Miao, Luyi and Xu, Yishu and Gao, Feng and Tian, Qi and Wang, Yizhou},
  title = {{Aligning Human Motion Generation with Human Perceptions}},
  booktitle = {ICLR},
  pages = {77626--77656},
  year = {2025}
}

@inproceedings{ling2025vmbench,
  author = {Ling, Xinran and Zhu, Chen and Wu, Meiqi and Li, Hangyu and Feng, Xiaokun and Yang, Cundian and Hao, Aiming and Zhu, Jiashu and Wu, Jiahong and Chu, Xiangxiang},
  title = {{VMBench: A Benchmark for Perception-Aligned Video Motion Generation}},
  booktitle = {ICCV},
  pages = {13087--13098},
  year = {2025}
}

@inproceedings{meng2024phygenbench,
  author = {Fanqing Meng and Jiaqi Liao and Xinyu Tan and Quanfeng Lu and Wenqi Shao and Kaipeng Zhang and Yu Cheng and Dianqi Li and Ping Luo},
  title = {{Towards World Simulator: Crafting Physical Commonsense-Based Benchmark for Video Generation}},
  booktitle = {ICML},
  pages = {43781--43806},
  year = {2025}
}

@inproceedings{motamed2025physicsiq,
  author = {Motamed, Saman and Culp, Laura and Swersky, Kevin and Jaini, Priyank and Geirhos, Robert},
  title = {{Do Generative Video Models Understand Physical Principles?}},
  booktitle = {WACV},
  pages = {948--958},
  year = {2026}
}

@article{cai2024temporalbench,
  author = {Cai, Mu and Tan, Reuben and Zhang, Jianrui and Zou, Bocheng and Zhang, Kai and Yao, Feng and Zhu, Fangrui and Gu, Jing and Zhong, Yiwu and Shang, Yuzhang and Dou, Yao and Park, Jaden and Gao, Jianfeng and Lee, Yong Jae and Yang, Jianwei},
  title = {{TemporalBench: Benchmarking Fine-grained Temporal Understanding for Multimodal Video Models}},
  journal = {arXiv preprint arXiv:2410.10818},
  year = {2024}
}

@article{zhang2024vinoground,
  author = {Zhang, Jianrui and Cai, Mu and Lee, Yong Jae},
  title = {{Vinoground: Scrutinizing LMMs over Dense Temporal Reasoning with Short Videos}},
  journal = {arXiv preprint arXiv:2410.02763},
  year = {2024}
}

@inproceedings{cores2024tvbench,
  author = {Daniel Cores and Michael Dorkenwald and Manuel Mucientes and Cees G. M. Snoek and Yuki M Asano},
  title = {{Lost in Time: A New Temporal Benchmark for VideoLLMs}},
  booktitle = {BMVC},
  year = {2025}
}

@inproceedings{xiong2024llavacritic,
  author = {Xiong, Tianyi and Wang, Xiyao and Guo, Dong and Ye, Qinghao and Fan, Haoqi and Gu, Quanquan and Huang, Heng and Li, Chunyuan},
  title = {{LLaVA-Critic: Learning to Evaluate Multimodal Models}},
  booktitle = {CVPR},
  pages = {13618--13628},
  year = {2025}
}

@article{wang2024selftaught,
  author = {Wang, Tianlu and Kulikov, Ilia and Golovneva, Olga and Yu, Ping and Yuan, Weizhe and Dwivedi-Yu, Jane and Pang, Richard Yuanzhe and Fazel-Zarandi, Maryam and Weston, Jason and Li, Xian},
  title = {{Self-Taught Evaluators}},
  journal = {arXiv preprint arXiv:2408.02666},
  year = {2024}
}

@article{du2026rsimaster,
  author = {Du, Yaxin and Yang, Xiyuan and Zhou, Zhifan and Ge, Yujie and Wang, Cheng and Wang, Jiajun and Chen, Sijie and Liu, Zehui and Zhang, Yuxin and Gu, Weicheng and Zhang, Julian and Lei, Zixing and Chen, Siheng},
  title = {{RSI-Master: Structuring Experiments to Guide Autonomous Model Improvement}},
  journal = {arXiv preprint arXiv:2609.35561},
  year = {2026}
}

@article{du2026datamaster,
  author = {Du, Yaxin and Yang, Xiyuan and Zhou, Zhifan and Liu, Wanxu and Lei, Zixing and Chen, Zimeng and Liu, Fenyi and Wu, Haotian and Cai, Yuzhu and Liu, Zexi and Zhu, Xinyu and Wang, WenHao and Zhang, Linfeng and Qian, Chen and Chen, Siheng},
  title = {{DataMaster: Data-Centric Autonomous AI Research}},
  journal = {arXiv preprint arXiv:2605.10906},
  year = {2026}
}

@inproceedings{farneback2003flow,
  author = {Gunnar Farneb{\"a}ck},
  title = {Two-Frame Motion Estimation Based on Polynomial Expansion},
  booktitle = {Proceedings of the 13th Scandinavian Conference on Image Analysis},
  pages = {363--370},
  year = {2003},
  series = {Lecture Notes in Computer Science},
  volume = {2749}
}
